\documentclass{article}

\usepackage[preprint]{neurips_2026}

\usepackage[utf8]{inputenc} %
\usepackage[T1]{fontenc}    %
\usepackage{hyperref}       %
\usepackage{url}            %
\usepackage{booktabs}       %
\usepackage{amsfonts}       %
\usepackage{nicefrac}       %
\usepackage{microtype}      %
\usepackage{xcolor}         %
\usepackage{graphicx}
\usepackage{subcaption}
\usepackage[normalem]{ulem}
\usepackage{amssymb}
\usepackage{amsmath}
\usepackage{multirow}
\usepackage{makecell}

\newcommand{\upscale}[1]{\hspace*{0.05cm}\!\uparrow\!\hspace*{0.05cm}#1}

\title{From Global to Local: Efficient Regional Weather Downscaling with Global Weather Foundation Model}

\author{%
  Wiktor Kamzela$^{1, 2}$ \And Jakub Kubiak$^{1,2}$ \And Adam Dobosz$^{1, 2}$ \And Jędrzej Miczke$^{1, 2}$ \And Anatol Kaczmarek$^{1, 2}$ \And Piotr Wyrwiński$^{1, 2}$ \And Wojciech Stefaniak$^{1}$ \And Wojciech Kotłowski$^{2}$ \\
  \\ %
  $^{1}$Poznań Supercomputing and Networking Center \\
  $^{2}$Poznań University of Technology \\
  \\ 
  \texttt{wstefaniak@man.poznan.pl} \\
}

\begin{document}

\maketitle

\begin{abstract}
Accurate regional weather prediction requires resolving fine-scale structure while remaining consistent with global dynamics. Traditional limited area models rely on computationally expensive simulations, while many learning-based approaches frame the problem as super-resolution, overlooking statistical and physical mismatches across scales.
We propose a foundation-model-driven downscaling framework that learns regional refinements of global forecasts by augmenting a pretrained weather model backbone with lightweight, multi-scale prediction heads operating directly in its latent space. Despite being trained on substantially coarser inputs, the pretrained backbone supports regional adaptation at resolutions corresponding to a two-order-of-magnitude increase in grid-cell resolution, without the need for retraining.  
The proposed approach uses regional numerical simulations as training targets and is evaluated not only against gridded datasets but also against ground-based weather station observations, enabling analysis of systematic biases between global reanalysis, regional simulations, and in-situ weather station observations. Our experiments show improved accuracy in comparison to NWP on most of the metrics at the fraction of computational cost. 
Moreover, we observe that building on a latent space of globally pre-trained weather foundation model offers better downscaling capabilities than the standard image-based super-resolution approaches. 
\end{abstract}

\section{Introduction}
Accurate weather prediction, normally solved by Numerical Weather Prediction (NWP) models \cite{NWP} became one of the key technologies in  today's applications spanning from precision farming \cite{NN_agri} and urban flood management \cite{flash_flooding} to stabilization of the energy grid by accurate predictions of renewable energy sources \cite{RES}. However, all of the aforementioned applications require forecasts that are not only precise but also produced quickly. NWP methods address this challenge through localized mesoscale approaches, such as the Weather Research and Forecasting (WRF) model \cite{WRF_ver3}, including its most recent state-of-the-art solvers: Advanced Research WRF (WRF-ARW) \cite{WRF-ARW} and the nonhydrostatic mesoscale model (WRF-NMM) \cite{WRF-NMM}. While the precision requirement is clearly met -- since NWP outputs have become standard inputs for Data-Driven Models (DDMs) -- the speed requirement remains a limitation due to the computationally intensive nature of NWP simulations \cite{vaughan2024}. This limitation has driven growing interest in alternative methods for weather prediction, such as DDMs.
Recently, AI models such as GraphCast, Aurora or FourCastNet \cite{graphcast, aurora, fourcastnet} have matched NWP accuracy, providing global weather forecasts at $0.25^\circ$ resolution, limited by the ERA5 dataset \cite{era5_2020}, while being orders of magnitude faster. Moreover, even subjects responsible for providing global forecasts using traditional methods, took interest in DDMs and created their own models \cite{AIFS}. On the same level, ML-based Limited Area Models (ML LAMs) became a trusted substitute of traditional mesoscale methods, aimed at modelling weather phenomena at finer scales, providing both accuracy and speed. However, until recently they heavily relied on input from NWP models for stabilizing training, by providing the most accurate Local Boundary Condition (LBC) \cite{YingLong, diffusionLAM}. This, and the fact that most of the ML LAM models need to be trained from scratch \cite{adamov2025} or fine-tuned \cite{munir2024} is the motivation of our work. 
In this work we propose a new approach for building weather downscaling models that utilizes pretrained foundation model -- Aurora \cite{aurora}, without the need of fine-tuning most of the architecture involving heavy computation. We use standard ERA5 $0.25^\circ$ input states at given timestamps to generate outputs at the WRF $0.025^\circ$ resolution for surface-level weather variables such as wind, temperature, and pressure. In addition, we evaluate our approach on all US-based HadISD \cite{HadISD} in-situ observations, including coastline stations and HRRR forecasts dataset produced by the WRF-ARW model. \cite{HRRR}.

Our main contributions are:
\begin{itemize}
    \item \textbf{LBC-Free Regional Downscaling:} We propose a framework that eliminates dependency on NWP-based Local Boundary Conditions (LBCs) for Limited Area Models. Unlike traditional ML-LAMs that require region-specific retraining and fail when driving regional forecasts are unavailable, our method uses global foundation model representations directly, improving operational robustness and simplifying the training pipeline.
    \item \textbf{Latent-Space Regional Adaptation:} We demonstrate that latent-space decoupling enhances adaptation from global to regional scales, allowing a frozen 0.25° model to perform 0.025° mesoscale forecasting without backbone retraining, using lightweight decoder heads.
    \item \textbf{Operational Benchmarking:} We evaluate against operational NWP (WRF-ARW) and global foundation baselines on both gridded and in-situ weather stations data, demonstrating improved accuracy at inference speeds orders of magnitude faster than NWP.
\end{itemize}

\begin{figure*}[!t]

    \centering
    \includegraphics[width=\linewidth]{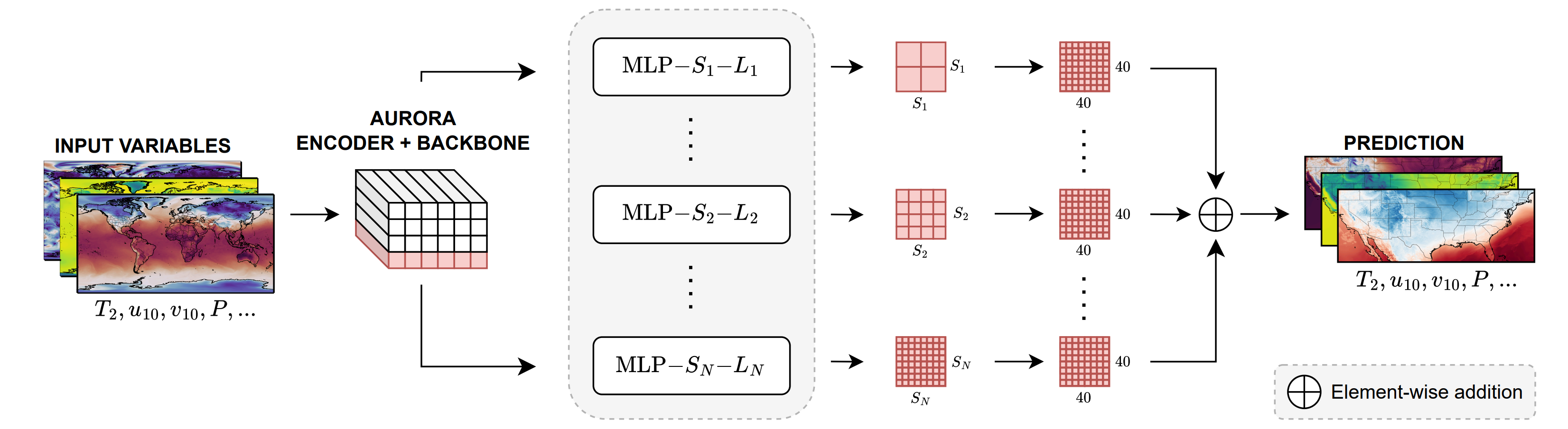}
    \caption{
        Overview of the workflow. Decoder variants augment or replace Aurora’s read-out with additional linear or MLP heads operating at different spatial resolutions. For simplicity, decoder heads are denoted as MLP-S-L, where $S$ indicates the native output resolution and $L$ the number of hidden layers. The special case $L=0$ corresponds to a linear decoder. In general, each head may use arbitrary $(S,L)$, and multiple heads ($n=1,\dots,N$) can be composed. The figure shows only the case of parallel heads with elementwise summation after upsampling to $40\times40$, while sequential compositions are omitted. The architecture is agnostic to the specific choice of input and target variables, subject to the constraints imposed by the pretrained backbone; the variables shown correspond to those used in our experiments.
    }
    \label{fig:model-architecture}
\end{figure*}

\section{Related Works}
Early machine learning approaches were explored as data-driven alternatives to numerical weather prediction (NWP) due to the high computational and memory costs of traditional models. Using historical reanalysis datasets such as ERA-Interim and ERA5, these methods framed weather forecasting as a supervised learning problem, learning mappings between atmospheric states and enabling short- to medium-range forecasts at significantly reduced inference cost \cite{dueben2018challenges}.

Convolutional neural networks (CNNs) were among the first deep learning models applied successfully to weather forecasting, owing to their ability to capture spatial correlations on structured grids. Early CNN-based models achieved promising performance for variables such as geopotential height, temperature, and wind, while offering substantial speed-up over numerical baselines inference \cite{weyn2019can}. However, their reliance on purely spatial convolutions and limited temporal modeling hindered their ability to capture long-range temporal dependencies and multiscale atmospheric dynamics, ultimately constraining forecast lead times and generalization \cite{weyn2020improving}.

To address the limitations of early CNN-based models, recent work has shifted toward large-scale deep learning systems that jointly model atmospheric dynamics across space and time using more advanced architectures such as GNNs and transformers \cite{GNN, Transformer}. These end-to-end approaches aim to match operational NWP skill at substantially lower inference cost.

Early successes include FourCastNet \cite{fourcastnet}, which applied Fourier Neural Operators to model global weather fields with competitive medium-range skill and large speedups over NWP. In contrast, Pangu-Weather used a hierarchical transformer to capture multiscale atmospheric structure and achieved state-of-the-art performance. \cite{pangu}. Subsequent models expanded this paradigm: GraphCast formulated global forecasting on a spherical graph for improved spatial modeling \cite{graphcast}, while ClimaX introduced a foundation model approach emphasizing scalability and adaptability across weather and climate tasks \cite{climax}. One of the most successful global models is Aurora \cite{aurora}, a foundation model trained on vast amounts of data, enabling it to learn a strong latent representation of the climate. We build on this work by adopting the $0.25^\circ$ resolution base model, which was originally designed for fine-tuning across a range of downstream tasks. Despite these advances, such models operate at relatively coarse spatial resolutions (e.g., $0.25^\circ$), limiting their applicability to mesoscale forecasting.

In recent years Machine Learning based Limited Area Models (ML LAMs) started to emerge as a solution for replacing mesoscale simulations, offering fine-scale predictions at a fraction of time consumed by NWP models. First ML LAMs required input from traditional forecasts as a way to stabilize training. They employ ground truth weather states as boundary forcing conditions \cite{oskarsson2023, YingLong} to keep errors from accumulating across next roll-out steps. The most recent work \cite{adamov2025} managed to bring ML LAM closer to practical scenarios by using ML-based forecasts as LBC. They needed, however, to include future boundary forcing i.e. a state of local boundary conditions from time $t+1$, making it easier for model to predict region of interest. In other approaches stretched-grid models were applied to combine global and local meshes of data to achieve promising results on regional forecasts \cite{nipen2024}. While all aforementioned methods solely depend on training new models rather than using existing ones, developed for coarser scale, \cite{munir2024} presented approach of fine-tuning pretrained foundation model ClimaX, effectively downscaling weather forecasts  $5.652^\circ \rightarrow 1.40652^\circ$, showing that fine-tuning can be leveraged to achieve better performance, but failing to demonstrate it on a more challenging fine-scaled grid.

\section{Dataset}
\textbf{Wind Toolkit US.} \label{sec:wtk}The Wind Toolkit–United States (WTK–US) \cite{WRF_dataset} is a high-resolution dataset generated using the Weather Research and Forecasting (WRF) model \cite{WRF_ver3}. It spans 2007–2013 and provides meteorological fields over the CONUS (CONtiguous United States) region at a 5-minute temporal resolution. The dataset is defined on a Lambert Conformal Conic (LCC) projected grid, which is uniform in projected space but curvilinear in geographic coordinates. In practice, grid cells are distributed approximately uniformly in physical distance (km), yielding near-uniform spatial density across the domain, in contrast to latitude-longitude grids ($^\circ$), where density increases toward the poles. The WRF model used to generate WTK--US is a numerical weather prediction system initialized from large-scale reanalysis data and produces outputs at approximately 2 km resolution. In our approach, it serves exclusively as the training target, against which the loss function is computed.

\textbf{High-Resolution Rapid Refresh (HRRR).} \label{sec:hrrr}
The High-Resolution Rapid Refresh (HRRR) dataset is a NWP product generated using the WRF-ARW modeling system with rapid-update data assimilation. It provides three-dimensional atmospheric fields over the CONUS domain at $\approx$3 km horizontal resolution and hourly temporal resolution, defined on a Lambert Conformal Conic (LCC) projected grid. Although the dataset starts in 2014, early years contain substantial monthly gaps, therefore, we use data from 2017–2024. The analysis fields are produced through frequent cycling data assimilation incorporating diverse observational inputs, yielding dynamically consistent estimates of the atmospheric state. In this work, HRRR serves a role analogous to WTK-US, except that 2022 is reserved for evaluation to enable direct comparison with the Aurora base model. In addition to assimilation data, the HRRR dataset includes accurate, state-of-the-art NWP forecasts produced using the WRF-ARW model, which we use to evaluate our data-driven approach. For simplicity we refer to the assimilation (+0h) data as HRRR and predictions (+6h-48h) generated with WRF-ARW model as WRF-ARW. The pressure experiments use HRRR data from 2017–2019 for training and 2020 for evaluation, excluding December due to a known issue affecting this variable between December 2020 and July 2022, documented in a public note \cite{NOAA2022_hrrr_great_lakes}. A detailed analysis and justification of the selected time range are provided in Appendix~\ref{appendix:hrrr_great_lakes}.

\textbf{ERA5.} \label{sec:era}
ERA5 ~\cite{store2023era5, hersbach2020era5} is the fifth-generation atmospheric reanalysis dataset spanning the period from 1940 to 2024, providing global atmospheric fields at hourly temporal resolution on a regular rectilinear grid with a spatial resolution of $0.25^\circ$ (approximately 28 km at the equator). The dataset stores a wide set of variables that can be grouped into 3 main categories: atmospheric state, near surface/surface variables, and static variables (e.g. land-sea mask or orography). The atmospheric state in ERA5 is reconstructed using four-dimensional variational data assimilation (4D-Var), which combines a numerical weather prediction model with temporally distributed observations to produce a physically consistent estimate of the atmospheric state. 
In our approach, ERA5 serves as input for extracting latent $1^\circ \times 1^\circ$ spatial embeddings from Aurora’s backbone, from which weather variables are predicted at higher spatial resolution.

\textbf{WTK--US and HRRR regridding.} \label{par:wtk_us_regridding}
Both WTK--US and HRRR datasets are originally defined using the LCC curvilinear projection, whereas the ERA5 dataset utilizes a $0.25^\circ$ latitude--longitude grid. For our analysis, the WTK--US and HRRR datasets are reprojected onto a regular latitude--longitude grid aligned with the ERA5 coordinate system, while preserving their native high spatial resolution.

The regridding process was carried out in a physically grounded manner, preserving both spatial and physical consistency for each variable. Intensive variables were treated pointwise using bilinear interpolation, extensive variables were regridded to conserve totals, and wind variables were decomposed into their orthogonal vector components. This careful, variable-aware handling of the regridding process is crucial for preserving meaningful fine-scale information without introducing spurious artifacts, and it ensures that the resulting dataset is suitable for super-resolution modelling.
Additional implementation details along with the calculation of two-step regridding error are provided in Appendix~\ref{appendix:regridding} and ~\ref{appendix:Regridding-roundtrip-error}. 

\textbf{HadISD.} \label{sec:hadisd}
The Hadley Integrated Surface Dataset (HadISD) \cite{HadISD} is a quality-controlled collection of in-situ surface meteorological observations derived from global weather station records. For this study, we focus on the subset of stations located within the CONUS region, comprising approximately 1,800 stations with irregular spatial distribution. HadISD provides point-based measurements of key near-surface variables and serves as an independent observational reference that is not influenced by model dynamics. In our approach, this dataset is used exclusively as a benchmark for evaluation, with all assessments conducted on data from the year 2022. The evaluation metric (RMSE) is computed between station measurements and nearest $0.025^\circ$ grid points, with station values averaged within grid cells for intensive variables.

\begin{figure*}[!t]
    \centering
    \includegraphics[width=0.85\linewidth]{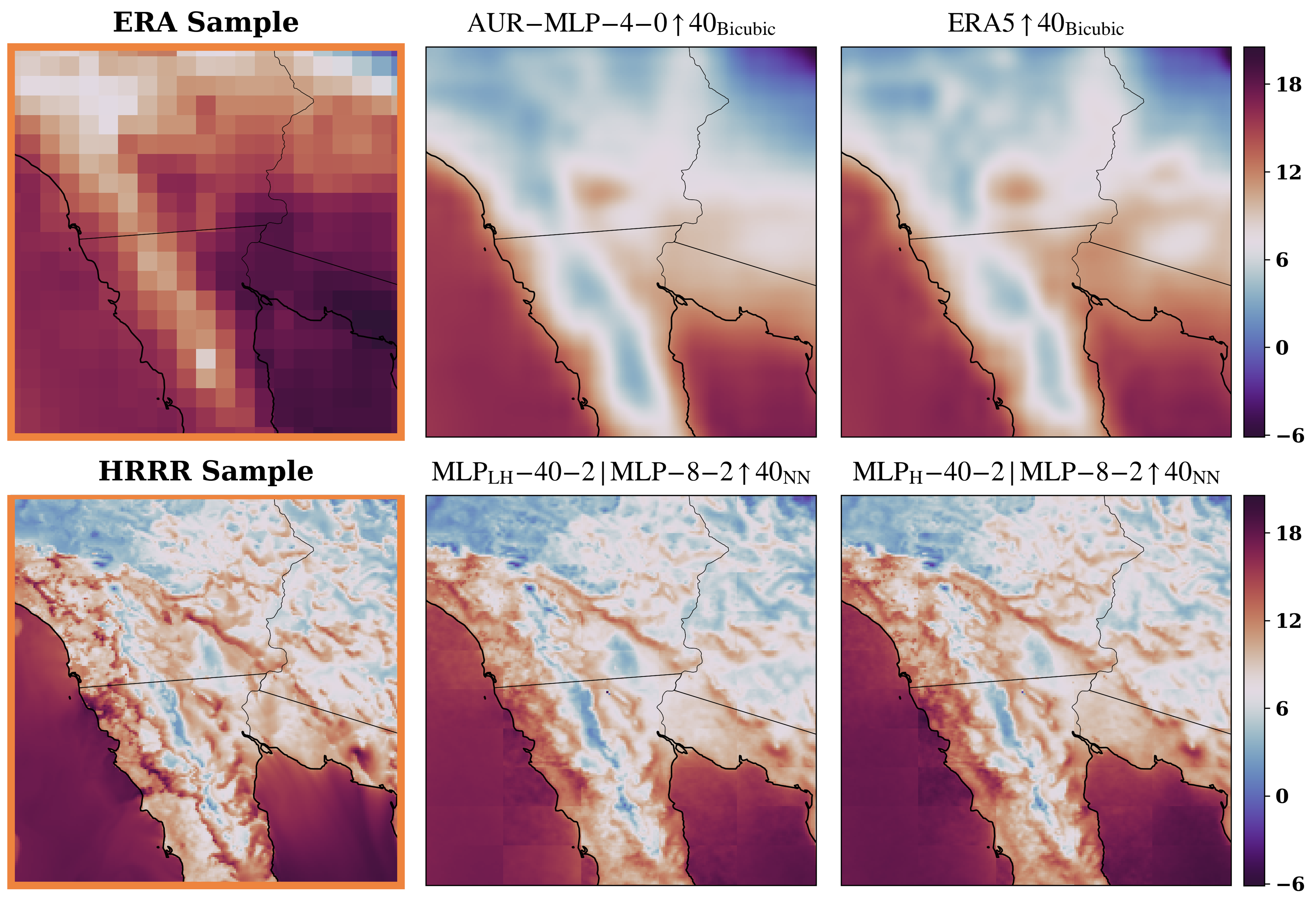}
    \caption{
        Comparison of predictions for a selected sub-domain of the CONUS region.
    }
    \label{fig:results_graphic}
\end{figure*}

\section{Method}
We build on the state-of-the-art Aurora foundation model \cite{aurora}, which learns generalizable representations of global weather and climate dynamics from diverse reanalysis, forecast, and climate data. Rather than modifying its pretrained backbone, we use Aurora as a fixed simulator and adapt its decoder to produce outputs at the desired spatial resolution and 6-hour temporal resolution. We extend the model with specialized output heads for each target variable, enabling each head to leverage shared features for domain-specific tasks.
Aurora encodes global atmospheric dependencies at the level of the backbone, where patch embeddings are computed jointly and contextualized through cross-patch interactions. As a result, each embedding represents a local spatial region conditioned on large-scale and non-local atmospheric structure learned during pretraining. In contrast, decoding in Aurora factorizes over patches: given these contextualized embeddings, outputs are produced independently for each patch. This separation between global coupling in the backbone and patch-local decoding not only allows decoder heads to be trained as lightweight, modular adapters without sacrificing access to global weather dynamics, but also allows to have equally accurate outputs in the whole prediction domain, without the drop of quality along boundaries.
In principle, this patch-local decoding interface also allows a single decoder head to be trained on patches drawn from multiple geographic regions or datasets. Such multi-regional training proves effective, with no substantial increase in loss. We provide examples of these training runs in Appendix~\ref{appendix:multiregional}. 

All models are trained and evaluated at a target resolution of $40\times40$ per patch. To study resolution-capacity trade-offs and reduce the number of trainable parameters, some decoder heads predict lower-resolution outputs that are deterministically upsampled to $40\times40$ using nearest-neighbor replication. This upsampling is non-learned and applied uniformly across methods.
We consider four types of decoder components, which can be composed into more complex models evaluated in our experiments~(Figure~\ref{fig:model-architecture}). 

\textbf{(1) Aurora linear decoder.}
Aurora provides a pretrained linear decoder that maps 1024-dimensional embeddings to $4\times4$ output patches. For the consistency, we denote this decoder as AUR-MLP-4-0 with $4\times4$ output dimensions and 0 hidden layers (linear network). When evaluated at the target resolution, its outputs are deterministically upsampled to $40\times40$ either by neareast neighbour replication of each pixel $10\times10$ times ($\mathrm{AUR\text{-}MLP\text{-}4\text{-}0 \upscale{40}_{NN}}$) or by bicubic interpolation ($\mathrm{AUR\text{-}MLP\text{-}4\text{-}0 \upscale{40}_{Bicubic}}$).

\textbf{(2) Embedding-based MLP decoders (MLP-S-L).}
We introduce a family of decoder heads that map Aurora's 1024-dimensional embeddings directly to an $S\times S$ output. An \textbf{MLP-S-L} decoder consists of $L$ hidden layers, each of width 1024, with GELU activation functions, followed by a linear projection to the output resolution. The special case $L=0$ corresponds to a purely linear projection and recovers linear decoder heads trained from scratch, denoted in out work as \textbf{MLP-S-0}.

\textbf{(3) Patch-based MLP decoders (MLP\textsubscript{P}-S-L).}
To show benefits of utilizing latent space of learned embeddings, we compare our method to direct super-resolution approach by introducing a variant of the MLP decoder that operates on spatial patches rather than the full embedding. In MLP\textsubscript{P}-S-L, the MLP takes as input the $4\times4$ output patch produced by Aurora’s pretrained linear decoder instead of the 1024-dimensional embedding. The patch is flattened and processed by an MLP with $L$ hidden layers, producing an $S\times S$ output. This variant can be interpreted as a lightweight, learnable post-processing module that refines Aurora’s native patch-level predictions while preserving the patch-local decoding structure.

\textbf{(4) Orography encoders \textsubscript{LH} and \textsubscript{H}}
Orography is one of the key factors that enable high-resolution forecasting in NWP models. Although Aurora’s backbone has access to orography information matching its input resolution, following NWP initialization workflow, we decided to provide high-resolution elevation map to the input of our decoders. We consider two variants for incorporating this information, where the notation refers to the full model (orography encoding plus decoder): (i) direct inclusion of the full $40\times40$ orography field per patch, denoted MLP\textsubscript{H}, and (ii) a learned projection via a shallow MLP with a single hidden layer of size 1024, which encodes the orography into a 1024-dimensional embedding, denoted MLP\textsubscript{LH}. In both cases we stack these vectors with embeddings from Aurora's backbone.

\textbf{Head compositions.}
Decoder heads are treated as modular components and can be combined to form more expressive architectures. We consider two composition patterns. In sequential composition ("+"), the output of one module is provided as input to the next. In parallel composition ("$|$"), multiple decoder heads operate on the same Aurora embedding in parallel, and their outputs are summed to produce the final prediction. Parallel composition enables different heads to capture complementary aspects of the target field, such as coarse structure and fine-scale details. For decoder heads with $\mathrm{S}<40$, outputs involved in parallel compositions are brought to a common $40\times40$ resolution via deterministic upsampling marked with $\mathrm{\upscale{40}_{NN}}$ prior to summation.

\textbf{Baselines.}
We first compare our approach against several baseline methods of non-learned interpolation. These include both nearest-neighbor interpolation of ERA5 ($\mathrm{ERA5\upscale{40}_{NN}}$), bicubic variant ($\mathrm{ERA5\upscale{40}_{Bicubic}}$) and Aurora’s pretrained linear decoder, the output of which was upscaled from $4\times4$ through bicubic interpolation, denoted $\mathrm{AUR\text{-}MLP\text{-}4\text{-}0\upscale{40}_{Bicubic}}$. The next set of baselines spans learned super-resolution approaches including CNN-based super-resolution models such as FSRCNN~\cite{dong2016accelerating} and Fast-SRGAN~\cite{ledig2017srgan, fastsrgan}. In addition, we construct a composite baseline that explicitly leverages Aurora’s low-resolution outputs. It is based on learned patch-wise mapping, where $\mathrm{MLP_P\text{-}40\text{-}1}$ transforms each $\mathrm{AUR\text{-}MLP\text{-}4\text{-}0}$ output patch directly into a $40\times40$ prediction, denoted $\mathrm{AUR\text{-}MLP\text{-}4\text{-}0} + \mathrm{MLP_P\text{-}40\text{-}1}$. The fourth learned baseline follows the same patch-wise mapping strategy but replaces the MLP with a linear decoder, yielding $\mathrm{AUR\text{-}MLP\text{-}4\text{-}0} + \mathrm{MLP_P\text{-}40\text{-}0}$. These baselines isolate the effects of deterministic upsampling, linear vs. non-linear mappings, and multi-scale decoder designs on top of Aurora’s representations.

Additionally, we compare our approach against established baselines evaluated over the same time period to ensure a fair comparison. These include the state-of-the-art WRF-ARW model, Aurora’s $0.25^\circ$ base model with its original decoder fine-tuned on the HRRR dataset to align data biases, and 5-member ensemble StormCast \cite{stormcast}, a state-of-the-art diffusion model for kilometer-scale convective dynamics that represents a fundamentally different learning paradigm, spanned on LCC grid. Together, these baselines provide a comprehensive comparison across diverse modelling approaches.

\textbf{Experimental setup.} All experiments were conducted on NVIDIA H100 GPUs. A single training run takes approximately 10 hours, while inference requires around 1 minute per year on the same hardware. Training was performed on a single GPU with batch size 32 fitting within memory. To reduce I/O overhead and memory footprint, we operate directly on intermediate embeddings produced by Aurora’s backbone instead of raw ERA fields, leveraging their compressed representation. Consequently, the main memory requirement comes from storing these embeddings in system memory, requiring approximately 350 GB of CPU RAM, which can be avoided via lazy loading from disk. The WTK-US and HRRR datasets are also lazily read in chunks using Xarray. GPU memory usage remains modest (~20 GB), and all computations are performed in single precision (FP32). Additional details on data scale, embedding caching, and hardware configuration are provided in Appendix~\ref{appendix:data-compute-setup}.

We evaluate our approach on both gridded and in-situ datasets to assess downscaling accuracy and generalization across complementary settings. ERA5 provides low-resolution inputs, while WTK-US and HRRR serves as high-resolution ground truth for training. Evaluation is performed at a +6h lead time on (i) the HRRR grid, enabling dense, spatially aligned comparisons, and (ii) HadISD in-situ station observations, which test robustness against real-world measurement sparsity and representativeness errors. Evaluation on both datasets is performed using RMSE for near-surface temperature ($T_2$), horizontal wind components at 10 meters ($u_{10}$, $v_{10}$), and surface pressure ($P$).
Aurora-based models (i.e., the original, non-fine-tuned versions) are evaluated for temperature and wind only, as Aurora predicts mean sea level pressure rather than surface pressure and therefore does not provide a compatible pressure output. We also investigate rollout performance of our models to WRF-ARW NWP model to assess the stability of the approach. The models are implemented in PyTorch and trained with AdamW using an initial learning rate of 0.001 and cosine annealing with a 1000-step warm-up and a minimum learning rate of 0.0001. We use a batch size of 32, a weight decay of $1\times10^{-7}$, and a mean absolute error (MAE) loss with early stopping. Following Aurora's implementation, each decoder is trained independently with its own MAE loss. We do not apply spatial loss weighting, as we focus on local regional forecasting where the domains are geographically close and latitudinal effects are limited. Hyperparameters were selected via coarse grid search to ensure stable training and competitive performance.

\section{Results}
\begin{table}[t]
\caption{
Performance evaluation comparison between HRRR grid and HadISD in-situ observations using RMSE at +6h lead time.
$T_2$ denotes 2-m air temperature ($^\circ$C), $u_{10}$ and $v_{10}$ wind component speeds (m/s), and $P$ surface pressure (hPa).\\
}
\label{tab:main-results}
\centering
\setlength{\tabcolsep}{4pt}
\renewcommand{\arraystretch}{1.0}
\resizebox{\linewidth}{!}{
\begin{tabular}{@{}lrrrrrrrr@{}}
\toprule
& \multicolumn{4}{c}{HRRR} & \multicolumn{4}{c}{HadISD} \\
\cmidrule(lr){2-5} \cmidrule(lr){6-9}
Model & \multicolumn{1}{c}{$T_2$} & \multicolumn{1}{c}{$u_{10}$} & \multicolumn{1}{c}{$v_{10}$} & \multicolumn{1}{c}{$P$} & \multicolumn{1}{c}{$T_2$} & \multicolumn{1}{c}{$u_{10}$} & \multicolumn{1}{c}{$v_{10}$} & \multicolumn{1}{c}{$P$} \\
\midrule
$\mathrm{ERA5\upscale{40}_{Bicubic}}$ & 2.3496 & 1.8581 & 1.8902 & 13.3972 & 1.7212 & 1.8112 & 1.8112 & 13.8708 \\
$\mathrm{ERA5\upscale{40}_{NN}}$ & 2.3734 & 1.8680 & 1.8994 & 14.0689 & 2.0306 & 1.8232 & 1.8232 & 14.3087 \\
$\mathrm{AUR\text{-}MLP\text{-}4\text{-}0\upscale{40}_{Bicubic}}$ & 2.2606 & 1.8209 & 1.8512 & $--^{*}$ & 1.9550 & 1.7037 & 1.7990 & $--^{*}$ \\
\midrule
$\mathrm{AUR\text{-}MLP\text{-}4\text{-}0\upscale{40}_{NN}}$ \footnotesize(HRRR biased) & 1.6397 & 1.5831 & 1.5782 & 12.6193 & 2.0066 & 1.6588 & 1.7261 & 12.6100 \\
$\mathrm{FSRCNN}$ & 3.9614 & 2.0776 & 2.3357 & 19.1048 & 5.8717 & 2.2053 & 2.4033 & 18.7897 \\
$\mathrm{Fast\text{-}SRGAN}$ & 2.3531 & 1.6900 & 1.7408 & 2.2926 & 2.8084 & 1.7912 & 1.8791 & 4.8471 \\
$\mathrm{AUR\text{-}MLP\text{-}4\text{-}0\upscale{40}_{NN}} + \mathrm{MLP_P\text{-}40\text{-}0}$ & 1.5912 & 1.5684 & 1.5668 & 11.5562 & 1.9823 & 1.6421 & 1.7125 & 11.5377 \\
$\mathrm{AUR\text{-}MLP\text{-}4\text{-}0\upscale{40}_{NN}} + \mathrm{MLP_P\text{-}40\text{-}1}$ & 1.5864 & 1.5107 & 1.5103 & 3.0022 & 2.0013 & 1.6285 & 1.6866 & 5.5510 \\
\midrule
$\mathrm{AUR\text{-}MLP\text{-}4\text{-}0\upscale{40}_{NN}}$ &  &  &  &  &  &  &  &  \\
\hspace*{0.4cm} $\mid \mathrm{MLP\text{-}40\text{-}0}$ & 1.5648 & 1.5465 & 1.5510 & 11.0277 & 1.9730 & 1.6353 & 1.7055 & 11.2287 \\
\hspace*{0.4cm} $\mid \mathrm{MLP\text{-}40\text{-}1}$ & 1.3859 & 1.3710 & 1.3950 & 0.9252 & 1.8571 & 1.5607 & 1.6196 & 4.5909 \\
\hspace*{0.4cm} $\mid \mathrm{MLP\text{-}40\text{-}0} \mid \mathrm{MLP\text{-}8\text{-}0\upscale{40}_{NN}}$ & 1.5707 & 1.5468 & 1.5476 & 11.0262 & 1.9883 & 1.6363 & 1.7038 & 11.2257 \\
\hspace*{0.4cm} $\mid \mathrm{MLP\text{-}40\text{-}1} \mid \mathrm{MLP\text{-}8\text{-}1\upscale{40}_{NN}}$ & 1.3277 & 1.3391 & 1.3606 & 0.7023 & 1.7964 & 1.5431 & 1.5998 & 4.5694 \\
\hspace*{0.4cm} $\mid \mathrm{MLP\text{-}40\text{-}2} \mid \mathrm{MLP\text{-}8\text{-}2\upscale{40}_{NN}}$ & 1.2870 & 1.2880 & 1.3128 & \textbf{0.4481} & 1.7427 & 1.5234 & 1.5779 & 4.3617 \\
\hspace*{0.4cm} $\mid \mathrm{MLP_{H}\text{-}40\text{-}2} \mid \mathrm{MLP\text{-}8\text{-}2\upscale{40}_{NN}}$ & 1.2865 & 1.2873 & 1.3185 & 0.5149 & 1.7475 & 1.5227 & 1.5801 & \textbf{4.3503} \\
\hspace*{0.4cm} $\mid \mathrm{MLP_{LH}\text{-}40\text{-}2} \mid \mathrm{MLP\text{-}8\text{-}2\upscale{40}_{NN}}$ & \textbf{1.2496} & \textbf{1.2758} & \textbf{1.3051} & 0.5140 & \textbf{1.6978} & \textbf{1.5180} & \textbf{1.5723} & 4.3670 \\
\bottomrule
\end{tabular}
}
\footnotesize
\parbox{\linewidth}{
\raggedright
$^*$ Aurora does not provide surface level pressure.
}
\end{table}

\begin{table}[t]
\caption{
Performance comparison between models with and without the pretrained Aurora linear decoder head ($\mathrm{AUR\text{-}MLP\text{-}4\text{-}0\upscale{40}}$).
Chosen only 2 well-performing models for comparison. Evaluation performed on the HRRR grid using RMSE at +6h lead time.
$T_2$ denotes 2-m air temperature ($^\circ$C), $W_{10}$ average over $U$ and $V$ wind speed components (m/s), and $P$ surface pressure (hPa).\\
}
\centering
\small
\begin{tabular}{@{}lcccc@{}}
\toprule
& \multirow[b]{2}{*}{\begin{tabular}{c}Pretrained\\decoder\end{tabular}} & \multicolumn{3}{c}{RMSE} \\
\cmidrule(lr){3-5}
Model &  & $T_2$ & $W_{10}$ & $P$ \\
\midrule
$\mathrm{MLP\text{-}40\text{-}2}$ & $\times$ & 1.3329 & 1.3307 & 0.5004 \\
$\mathrm{MLP\text{-}40\text{-}2}$ & $\checkmark$ & \textbf{1.3197} & \textbf{1.3219} &  \textbf{0.4800} \\
\midrule
$\mathrm{MLP\text{-}40\text{-}2} \mid \mathrm{MLP\text{-}8\text{-}2\upscale{40}_{NN}}$ & $\times$ & 1.3015 & 1.3089 & \textbf{0.4421} \\
$\mathrm{MLP\text{-}40\text{-}2} \mid \mathrm{MLP\text{-}8\text{-}2\upscale{40}_{NN}}$ & $\checkmark$ & \textbf{1.2870} & \textbf{1.3004} & 0.4481 \\

\bottomrule
\end{tabular}
\label{tab:org-dec-comparison}
\end{table}

\begin{table}[t]
\caption{
Performance evaluation of our approach against WRF-ARW, StormCast 5-member Ensemble and Aurora’s $0.25^\circ$ base model. Results are presented for HadISD in-situ observations and the HRRR grid using RMSE at +6h lead time. 
$T_2$ denotes 2-m air temperature ($^\circ$C), $u_{10}$ and $v_{10}$ wind component speeds (m/s), and $P$ surface pressure (hPa).
\\}
\label{tab:wrf-arw}
\centering
\small
\setlength{\tabcolsep}{3pt}
\renewcommand{\arraystretch}{1.1}
\resizebox{0.95\linewidth}{!}{
\begin{tabular}{@{}lcccccccc@{}}
\toprule
& \multicolumn{4}{c}{HadISD (in-situ)} & \multicolumn{4}{c}{HRRR Grid} \\
\cmidrule(lr){2-5} \cmidrule(lr){6-9}
Model & $T_2$  & $u_{10}$ & $v_{10}$  & $P$ & $T_2$  & $u_{10} $ & $v_{10} $ & $P $ \\
\midrule
$\mathrm{AUR\text{-}MLP\text{-}4\text{-}0\upscale{40}_{NN}}$ &  &  &  &  &  &  &  &  \\
\hspace*{0.4cm} $\mid \mathrm{MLP_{LH}\text{-}40\text{-}2} \mid \mathrm{MLP\text{-}8\text{-}2\upscale{40}_{NN}}$ & \textbf{1.6978} & \textbf{1.5180} & \textbf{1.5723} & 4.3670 & 1.2496 & \textbf{1.2758} & \textbf{1.3051} & \textbf{0.5140} \\
WRF-ARW                 & 1.8832 & 1.7663 & 1.8229 & \textbf{4.3610} & \textbf{1.1749} & 1.3784 & 1.4248 & 0.7475 \\
Aurora $0.25^\circ$     & 2.0519 & 1.7290 & 1.8169 & --* & 1.8596 & 1.7671 & 1.7199 & --* \\
StormCast 5-member Ensemble     & 2.4279 & 1.8346 & 1.8939 & --** & 2.2355 & 1.5715 & 1.6308 & --** \\
\bottomrule
\end{tabular}%
}
\footnotesize
\parbox{\linewidth}{
\raggedright
$^*$ Aurora does not provide surface level pressure. $^{**}$ StormCast was trained on year 2020, which makes the comparison on surface pressure unfeasible.
}
\end{table}

Table~\ref{tab:main-results} summarizes the results for our defined methods. \textbf{Non-learned interpolation baselines} ($\mathrm{ERA5\upscale{40}_{NN}}$, $\mathrm{ERA5\upscale{40}_{Bicubic}}$) and $\mathrm{AUR\text{-}MLP\text{-}4\text{-}0\upscale{40}_{Bicubic}}$ as a group exhibit one of the largest errors across all variables, confirming that purely geometric upscaling is insufficient to recover fine-scale atmospheric structure.

\textbf{Learned super-resolution approaches} display heterogeneous performance that is strongly tied to architectural choices. The FSRCNN model consistently underperforms across all variables, indicating that its exclusive focus on spatial correlation is inadequate for capturing the temporally rich dynamics of atmospheric fields. In contrast, Fast‑SRGAN narrows the gap with MLP-based families, delivering a noticeable improvement over pure convolutional architecture. The most accurate results are achieved by custom decoder compositions that incorporate a pretrained Aurora decoder ($\mathrm{AUR\text{-}MLP\text{-}4\text{-}0}$) coupled with varying up‑sampling modules. Fine-tuned (i.e. de-biased) version of $\mathrm{AUR\text{-}MLP\text{-}4\text{-}0\upscale{40}_{NN}}$ beats its original counterpart improving on all variables and showing how dataset bias can influence overall scores. Both learned patch‑based ($\mathrm{MLP_P\text{-}40\text{-}0}$, $\mathrm{MLP_P\text{-}40\text{-}1}$) methods further decrease errors, due to additional learned projections, outperforming standard nearest‑neighbor up-sampler. Moreover, performance scales positively with depth, as deeper decoder stacks systematically reduce RMSE. Nevertheless, they still significantly lag behind architectures fed with embedding information extracted from Aurora's backbone.

\textbf{Embedding based models} compared to models operating directly on patched images ($\mathrm{MLP_P\text{-}40\text{-}L}$ vs. $\mathrm{MLP\text{-}40\text{-}L}$) show small improvement for linear model and clear advantage of embedding based model for MLP variant. Shallow heads ($\mathrm{MLP\text{-}40\text{-}0}$) perform similarly to the best learned baselines, while even the simplest mlp-based model ($\mathrm{MLP\text{-}40\text{-}1}$) certainly improves all scores on all variables.

\textbf{Multi-head architectures} consistently achieve the strongest performance. Parallel compositions of multi-layer heads outperform their single-head counterparts across all variables, confirming that distributing model capacity across spatial scales is more effective than relying on a single decoder. Increasing head depth improves performance, though with diminishing returns. Deeper heads substantially reduce pressure errors, while shallow models remain ineffective. The best-performing configuration without orography, $\mathrm{MLP\text{-}40\text{-}2} \mid \mathrm{MLP\text{-}8\text{-}2}$, achieves the lowest RMSE for temperature, both wind components and pressure. Incorporating MLP-encoded orography into the decoder further improves temperature and wind performance, while slightly degrading pressure. In contrast, directly stacking raw orography with the same architecture yields no improvement, indicating that even a lightweight learned projection of orographic features is more effective than using raw inputs.  In general, across all investigated architectures incorporating a lightweight, pre-trained decoder into our architecture yields consistent improvements (Table~\ref{tab:org-dec-comparison}). $\mathrm{AUR\text{-}MLP\text{-}4\text{-}0\upscale{40}_{NN}}$ is a small architecture that does not require much computational overhead, while enhancing overall scores and allowing faster convergence through better training initialization.

\textbf{Results on HadISD in-situ observations} (Table~\ref{tab:main-results}) closely mirror trends observed on the HRRR grid, despite the increased difficulty of sparse and potentially noisy measurements. Multi-head MLP architectures again outperform single-head variants, with $\mathrm{MLP_{LH}\text{-}40\text{-}2} \mid \mathrm{MLP\text{-}8\text{-}2\upscale{40}_{NN}}$ achieving the lowest RMSE for temperature and wind components. 
Importantly, the relative ranking of models is largely preserved between gridded and in-situ evaluations, indicating that gains are not specific to a particular grid or interpolation scheme. One of the outliers is the $\mathrm{ERA5\upscale{40}_{Bicubic}}$ model, which shows remarkably strong performance on the HadISD dataset, suggesting that sparse station coverage and spatial averaging effects can make coarse-resolution fields appear more accurate than expected at pointwise observation locations. More information on the error distribution of our best model on HadISD can be found in the appendix \ref{appendix:in-situ}.

\textbf{Comparison with other methods} is shown in Table~\ref{tab:wrf-arw}, which compare our strongest configuration against WRF-ARW, Stormcast 5-member Ensemble and Aurora's $0.25^\circ$ base model.  Consequently, Aurora is excluded from pressure comparisons on both HRRR and HadISD as it does not provide surface pressure output. On HadISD, our approach outperforms WRF-ARW across temperature and both wind components and is on par for pressure. Both StormCast-Ensemble and Aurora exhibits higher errors for all variables. On the HRRR grid, WRF-ARW attains the lowest RMSE for temperature, likely due to our model having inherited bias from the supplementary WTK-US dataset (see Appendix~\ref{appendix:data-bias}). However, our model outperforms WRF-ARW for both wind components and substantially for surface pressure, while also significantly improving upon Aurora across all comparable variables. According to Appendix~\ref{appendix:StormCast} we do not see results degradation caused by regridding, but to enable fair comparison we present StormCast results on LCC grid. Appendix~\ref{appendix:spectra} shows spectra analysis, indicating preservation of physical properties inherited from global, pretrained Aurora's backbone up to it's effective resolution, whereas behind that point we maintain close resemblance to NWP-based solvers. Moreover, Appendix~\ref{appendix:scaling-laws} shows that the approach scales well with data.

\begin{figure*}[!t]
    \centering
    \includegraphics[width=\linewidth]{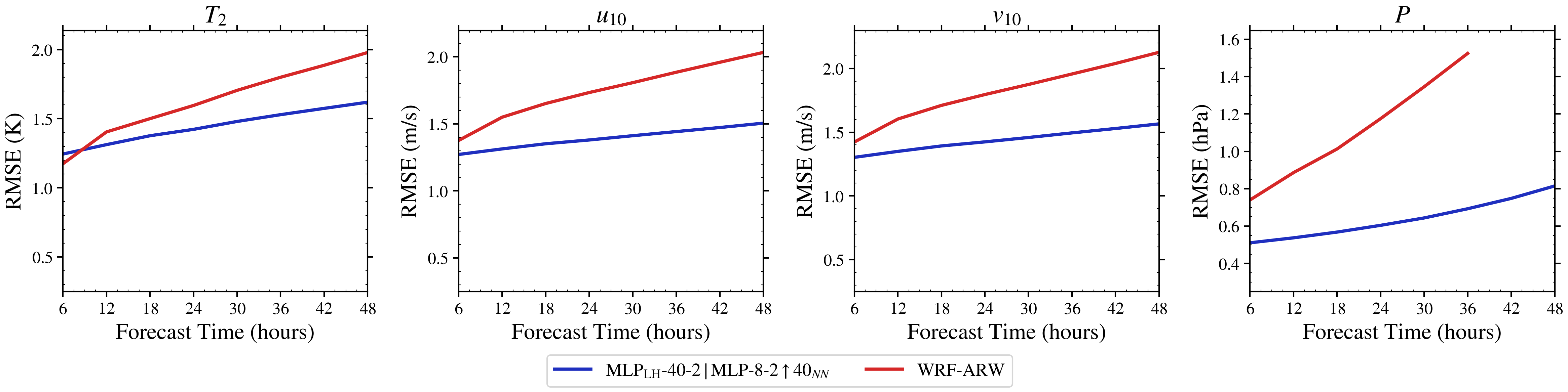}
    \caption{
        Performance comparison over a 48-hour horizon for the WRF-ARW model and our approach.
    }
    \label{fig:rollout}
\end{figure*}

\textbf{Rollout results} are shown in Figure~\ref{fig:rollout}, comparing our model against WRF-ARW predictions over a 48-hour horizon. Results are averaged over the full year 2022 for temperature and both wind components, and over 2020 for pressure. Since WRF-ARW forecasts for 2020 are not available up to the full 48-hour lead time, performance for this model is reported only up to 36 hours. For $T_{2m}$, our model is slightly worse at +6h but surpasses WRF-ARW from +12h onward across all remaining lead times. For $u_{10}$ and $v_{10}$, the results indicate improved stability of our approach, with an initially small advantage that becomes increasingly pronounced over time. Pressure shows a consistent advantage across the entire forecast horizon, with substantially reduced degradation. Overall, the average error increase over 48 hours is 18.5{\%} for our model compared to 75.5{\%} for WRF-ARW, indicating improved rollout stability over extended lead times. The largest difference is observed for pressure, where WRF-ARW shows over 130{\%} error growth despite a similar initial RMSE.

\section{Limitations}
Learning-based weather models inherit limitations from the data and simulations they are trained on, including systematic biases and uncertainties. As shown in this work, such biases can affect different variables and regions in non-uniform ways. In particular, careful examination of the HRRR dataset revealed inconsistencies that may influence model behavior and highlight the importance of scrutinizing training and evaluation data. Furthermore, our approach builds on a pre-trained foundation model backbone that captures multi-scale physical dynamics, however, the proposed heads operate on its latent representations without explicitly enforcing physical constraints. Even though we empirically verify that key properties, such as energy spectra, remain consistent, outputs produced by the proposed approach should therefore be interpreted as complementary to, rather than replacements for, established numerical weather prediction systems and expert analysis.

\section{Conclusion and future directions}
The proposed approach builds on Aurora’s foundation model backbone, which is designed to capture multi-scale physical structure, and combines it with simple downscaling heads to produce high-resolution outputs. Our results show an advantage of this approach over the commonly used WRF-ARW NWP model, defined baselines and StormCast. In addition, the use of a frozen backbone makes training efficient, requiring less than 10 hours on a single H100 GPU, enabling fast inference. Future work includes (1) validating model generalization beyond the training domain using diverse datasets, and (2) enhancing the architecture with multi-scale attention heads for improved spatial reasoning. (3) We also plan to extend Aurora’s design to include precipitation prediction, omitted in the original model. Leveraging high-resolution WTK-US and HRRR data, our method improves the precision and reliability of weather forecasts, supporting better decision-making in agriculture, disaster management, and renewable energy.

\bibliographystyle{unsrt}
\bibliography{bibliography}

\begin{thebibliography}{10}

\bibitem{NWP}
Weather prediction by numerical process. by {L}ewis {F}. {R}ichardson. cambridge (university press), 1922. 4°. pp. xii + 236. 30s.net.
\newblock {\em Quarterly Journal of the Royal Meteorological Society}, 48(203):282--284, 1922.

\bibitem{NN_agri}
Holger~R. Maier and Graeme~C. Dandy.
\newblock Neural networks for the prediction and forecasting of water resources variables: a review of modelling issues and applications.
\newblock {\em Environmental Modelling \& Software}, 15(1):101--124, 2000.

\bibitem{flash_flooding}
K.~Papagiannaki, K.~Lagouvardos, V.~Kotroni, and A.~Bezes.
\newblock Flash flood occurrence and relation to the rainfall hazard in a highly urbanized area.
\newblock {\em Natural Hazards and Earth System Sciences}, 15(8):1859--1871, 2015.

\bibitem{RES}
Kingsley Ukoba, Kehinde~O. Olatunji, Eyitayo Adeoye, Tien-Chien Jen, and Daniel~M. Madyira.
\newblock Optimizing renewable energy systems through artificial intelligence: Review and future prospects.
\newblock {\em Energy \& Environment}, 35(7):3833--3879, 2024.

\bibitem{WRF_ver3}
William Skamarock, Joseph Klemp, Jimy Dudhia, David Gill, Dale Barker, Michael Duda, Xiang-Yu Huang, Wei Wang, and Jordan Powers.
\newblock A description of the advanced research wrf version 3, 06 2008.

\bibitem{WRF-ARW}
William~C. Skamarock, Joseph~B. Klemp, Jimy Dudhia, David~O. Gill, Zhiquan Liu, Judith Berner, Wei Wang, John~G. Powers, Michael~G. Duda, Dale~M. Barker, and Xiang-Yu Huang.
\newblock A description of the advanced research wrf version 4.
\newblock Technical Report NCAR/TN-556+STR, National Center for Atmospheric Research, 2019.

\bibitem{WRF-NMM}
Zavi{\v s}a~I. Janjic.
\newblock A nonhydrostatic mesoscale model. part i: Dynamics.
\newblock {\em Meteorology and Atmospheric Physics}, 82(1--4):147--169, 2003.

\bibitem{vaughan2024}
Anna Vaughan, Stratis Markou, Will Tebbutt, James Requeima, Wessel~P. Bruinsma, Tom~R. Andersson, Michael Herzog, Nicholas~D. Lane, Matthew Chantry, J.~Scott Hosking, and Richard~E. Turner.
\newblock Aardvark weather: end-to-end data-driven weather forecasting, 2024.

\bibitem{graphcast}
Remi Lam, Alvaro Sanchez-Gonzalez, Matthew Willson, Peter Wirnsberger, Meire Fortunato, Ferran Alet, Suman Ravuri, Timo Ewalds, Zach Eaton-Rosen, Weihua Hu, Alexander Merose, Stephan Hoyer, George Holland, Oriol Vinyals, Jacklynn Stott, Alexander Pritzel, Shakir Mohamed, and Peter Battaglia.
\newblock Learning skillful medium-range global weather forecasting.
\newblock {\em Science}, 382(6677):1416--1421, 2023.

\bibitem{aurora}
Cristian Bodnar, Wessel~P Bruinsma, Ana Lucic, Megan Stanley, Anna Allen, Johannes Brandstetter, Patrick Garvan, Maik Riechert, Jonathan~A Weyn, Haiyu Dong, et~al.
\newblock A foundation model for the earth system.
\newblock https://github.com/microsoft/aurora/tree/main/aurora, 2025.
\newblock Released under the MIT License.

\bibitem{fourcastnet}
Jaideep Pathak, Shashank Subramanian, Peter Harrington, Sanjeev Raja, Ashesh Chattopadhyay, Morteza Mardani, Thorsten Kurth, David Hall, Zongyi Li, Kamyar Azizzadenesheli, et~al.
\newblock Fourcastnet: A global data-driven high-resolution weather model using adaptive fourier neural operators.
\newblock {\em arXiv preprint arXiv:2202.11214}, 2022.

\bibitem{era5_2020}
Hans Hersbach, Bill Bell, Paul Berrisford, Shoji Hirahara, András Horányi, Joaquín Muñoz-Sabater, Julien Nicolas, Caroline Peubey, Raluca Radu, Dinand Schepers, Adrian Simmons, Claudia Soci, Saleh Abdalla, Xavier Abellan, Gianpaolo Balsamo, Peter Bechtold, Giovanna Biavati, Jean-Raymond Bidlot, Massimo Bonavita, Giovanna De~Chiara, Per Dahlgren, Dick Dee, Michail Diamantakis, Rossana Dragani, Johannes Flemming, Richard Forbes, Manuel Fuentes, Alan Geer, Leopold Haimberger, Sean Healy, Robin~J. Hogan, Elías Hólm, Marta Janisková, Sarah Keeley, Patrick Laloyaux, Philippe Lopez, Cristian Lupu, Gabor Radnoti, Patricia de~Rosnay, Iryna Rozum, Freja Vamborg, Sebastien Villaume, and Jean-Noël Thépaut.
\newblock The era5 global reanalysis.
\newblock https://www.ecmwf.int/en/forecasts/dataset/ecmwf-reanalysis-v5, 2020.
\newblock Accessed via the Copernicus Climate Data Store, Released under Creative Commons Attribution-ShareAlike 4.0 International (CC BY-SA 4.0).

\bibitem{AIFS}
Simon Lang, Mihai Alexe, Matthew Chantry, Jesper Dramsch, Florian Pinault, Baudouin Raoult, Mariana C.~A. Clare, Christian Lessig, Michael Maier-Gerber, Linus Magnusson, Zied~Ben Bouallègue, Ana~Prieto Nemesio, Peter~D. Dueben, Andrew Brown, Florian Pappenberger, and Florence Rabier.
\newblock Aifs -- ecmwf's data-driven forecasting system, 2024.

\bibitem{YingLong}
Pengbo Xu, Xiaogu Zheng, Tianyan Gao, Yu~Wang, Junping Yin, Juan Zhang, Xuanze Zhang, San Luo, Zhonglei Wang, Zhimin Zhang, Xiaoguang Hu, and Xiaoxu Chen.
\newblock Yinglong-weather: Ai-based limited area models for forecasting of non-precipitation surface meteorological variables, 2025.

\bibitem{diffusionLAM}
Erik Larsson, Joel Oskarsson, Tomas Landelius, and Fredrik Lindsten.
\newblock Diffusion-lam: Probabilistic limited area weather forecasting with diffusion, 2025.

\bibitem{adamov2025}
Simon Adamov, Joel Oskarsson, Leif Denby, Tomas Landelius, Kasper Hintz, Simon Christiansen, Irene Schicker, Carlos Osuna, Fredrik Lindsten, Oliver Fuhrer, and Sebastian Schemm.
\newblock Building machine learning limited area models: Kilometer-scale weather forecasting in realistic settings, 2025.

\bibitem{munir2024}
Muhammad~Akhtar Munir, Fahad~Shahbaz Khan, and Salman Khan.
\newblock Efficient localized adaptation of neural weather forecasting: A case study in the {MENA} region, 2024.

\bibitem{HadISD}
R.~J.~H. Dunn, K.~M. Willett, P.~W. Thorne, E.~V. Woolley, I.~Durre, A.~Dai, D.~E. Parker, and R.~S. Vose.
\newblock Hadisd: a quality-controlled global synoptic report database for selected variables at long-term stations from 1973–2011.
\newblock {\em Climate of the Past}, 8(5):1649–1679, October 2012.
\newblock Data are available without charge under the UK Met Office Non-Commercial Government Licence.

\bibitem{HRRR}
David~C. Dowell, Curtis~R. Alexander, Eric~P. James, Stephen~S. Weygandt, Stanley~G. Benjamin, Geoffrey~S. Manikin, Benjamin~T. Blake, John~M. Brown, Joseph~B. Olson, Ming Hu, Tatiana~G. Smirnova, Terra Ladwig, Jaymes~S. Kenyon, Ravan Ahmadov, David~D. Turner, Jeffrey~D. Duda, and Trevor~I. Alcott.
\newblock The high-resolution rapid refresh (hrrr): An hourly updating convection-allowing forecast model. part i: Motivation and system description.
\newblock {\em Weather and Forecasting}, 37(8):1371--1395, 2022.
\newblock Available as part of the AWS Open Data Program. Data are in the public domain; no copyright restrictions. Users should provide appropriate attribution to NOAA.

\bibitem{dueben2018challenges}
Peter~D Dueben and Peter Bauer.
\newblock Challenges and design choices for global weather and climate models based on machine learning.
\newblock {\em Geoscientific Model Development}, 11(10):3999--4009, 2018.

\bibitem{weyn2019can}
Jonathan~A Weyn, Dale~R Durran, and Rich Caruana.
\newblock Can machines learn to predict weather? using deep learning to predict gridded 500-hpa geopotential height from historical weather data.
\newblock {\em Journal of Advances in Modeling Earth Systems}, 11(8):2680--2693, 2019.

\bibitem{weyn2020improving}
Jonathan~A Weyn, Dale~R Durran, and Rich Caruana.
\newblock Improving data-driven global weather prediction using deep convolutional neural networks on a cubed sphere.
\newblock {\em Journal of Advances in Modeling Earth Systems}, 12(9):e2020MS002109, 2020.

\bibitem{GNN}
Francesco Scarselli, Marco Gori, Ah~Chung Tsoi, Mauro Hagenbuchner, and Gabriele Monfardini.
\newblock The graph neural network model.
\newblock {\em IEEE Transactions on Neural Networks}, 2008.

\bibitem{Transformer}
Ashish Vaswani, Noam Shazeer, Niki Parmar, Jakob Uszkoreit, Llion Jones, Aidan~N Gomez, {\L}ukasz Kaiser, and Illia Polosukhin.
\newblock Attention is all you need.
\newblock In {\em Advances in Neural Information Processing Systems}, 2017.

\bibitem{pangu}
Kaifeng Bi, Lingxi Xie, Hengheng Zhang, Xin Chen, Xiaotao Gu, and Qi~Tian.
\newblock Pangu-weather: A 3d high-resolution model for fast and accurate global weather forecast.
\newblock {\em arXiv preprint arXiv:2211.02556}, 2022.

\bibitem{climax}
Tung Nguyen, Johannes Brandstetter, Ashish Kapoor, Jayesh~K Gupta, and Aditya Grover.
\newblock Climax: A foundation model for weather and climate.
\newblock {\em arXiv preprint arXiv:2301.10343}, 2023.

\bibitem{oskarsson2023}
Joel Oskarsson, Tomas Landelius, and Fredrik Lindsten.
\newblock Graph-based neural weather prediction for limited area modeling, 2023.

\bibitem{nipen2024}
Thomas~Nils Nipen, Håvard~Homleid Haugen, Magnus~Sikora Ingstad, Even~Marius Nordhagen, Aram Farhad~Shafiq Salihi, Paulina Tedesco, Ivar~Ambjørn Seierstad, Jørn Kristiansen, Simon Lang, Mihai Alexe, Jesper Dramsch, Baudouin Raoult, Gert Mertes, and Matthew Chantry.
\newblock Regional data-driven weather modeling with a global stretched-grid, 2024.

\bibitem{WRF_dataset}
Caroline Draxl, Brian Hodge, James McCaa, and Andrew Clifton.
\newblock Overview and meteorological validation of the wind integration national dataset toolkit.
\newblock https://www.nrel.gov/grid/wind-toolkit, 2015.
\newblock Released under Creative Commons Attribution 3.0 United States (CC BY 3.0 US).

\bibitem{NOAA2022_hrrr_great_lakes}
{National Weather Service} and {NCEP Environmental Modeling Center}.
\newblock Service change notice 22-68: Correction of the specification of elevation of the great lakes in the high resolution rapid refresh (hrrr).
\newblock \url{https://www.weather.gov/media/notification/pdf2/scn22-68_hrrr_great_lakes.pdf}, 2022.
\newblock Issued July 19, 2022.

\bibitem{store2023era5}
Copernicus Climate~Data Store et~al.
\newblock Era5 hourly data on single levels from 1940 to present.
\newblock {\em Copernicus Climate Change Service (C3S)}, 2023.

\bibitem{hersbach2020era5}
Hans Hersbach, Bill Bell, Paul Berrisford, Shoji Hirahara, Andr{\'a}s Hor{\'a}nyi, Joaqu{\'\i}n Mu{\~n}oz-Sabater, Julien Nicolas, Carole Peubey, Raluca Radu, Dinand Schepers, et~al.
\newblock The era5 global reanalysis.
\newblock {\em Quarterly journal of the royal meteorological society}, 146(730):1999--2049, 2020.

\bibitem{dong2016accelerating}
Chao Dong, Chen~Change Loy, and Xiaoou Tang.
\newblock Accelerating the super-resolution convolutional neural network.
\newblock In {\em European conference on computer vision}, pages 391--407. Springer, 2016.

\bibitem{ledig2017srgan}
Christian Ledig, Lucas Theis, Ferenc Husz{\'a}r, Jose Caballero, Andrew Cunningham, Alejandro Acosta, Andrew Aitken, Alykhan Tejani, Johannes Totz, Zehan Wang, et~al.
\newblock Photo-realistic single image super-resolution using a generative adversarial network.
\newblock In {\em Proceedings of the IEEE conference on computer vision and pattern recognition}, pages 4681--4690, 2017.

\bibitem{fastsrgan}
Hasnain Raz.
\newblock Fast-srgan.
\newblock \url{https://github.com/HasnainRaz/Fast-SRGAN}, 2019.
\newblock Accessed: 19 January 2026.

\bibitem{stormcast}
Jaideep Pathak, Yair Cohen, Piyush Garg, Peter Harrington, Noah Brenowitz, Dale Durran, Morteza Mardani, Arash Vahdat, Shaoming Xu, Karthik Kashinath, and Michael Pritchard.
\newblock Kilometer-scale convection allowing model emulation using generative diffusion modeling, 2024.

\bibitem{schluessel2007metop}
{Schl{\"u}ssel, Peter}.
\newblock First lessons learnt from metop.
\newblock \url{https://www.ecmwf.int/sites/default/files/elibrary/2007/15627-first-lessons-learnt-metop.pdf}, September 2007.
\newblock Presentation slides, 3--7 September 2007; Accessed: 17 January 2026.

\bibitem{era5_observations}
{ECMWF}.
\newblock Era5: Data documentation -- observations.
\newblock \url{https://confluence.ecmwf.int/display/CKB/ERA5%3A+data+documentation#ERA5:datadocumentation-Observations}, 2025.
\newblock Accessed: 17 January 2026.

\bibitem{esa_metop_handover2012}
{European Space Agency (ESA)}.
\newblock Esa hands over control of the metop-b weather satellite to eumetsat.
\newblock \url{https://www.esa.int/Applications/Observing_the_Earth/Meteorological_missions/MetOp/ESA_hands_over_control_of_the_Metop-B_weather_satellite_to_EUMETSAT}, September 2012.
\newblock Published: 21 September 2012; Accessed: 17 January 2026.

\bibitem{Danra}
Xiaohua Yang, Carlos Peralta, Bjarne Amstrup, Kasper~Stener Hintz, Søren~Borg Thorsen, Leif Denby, Simon~Kamuk Christiansen, Hauke Schulz, Sebastian Pelt, and Mathias Schreiner.
\newblock Danra: The kilometer-scale danish regional atmospheric reanalysis, 2025.
\newblock Dataset distributed under CC BY 4.0 license.

\end{thebibliography}

\appendix

\newpage
\section{Regridding - detailed description}\label{appendix:regridding}
\begin{figure}[!h]
    \centering
    \includegraphics[width=0.7\linewidth]{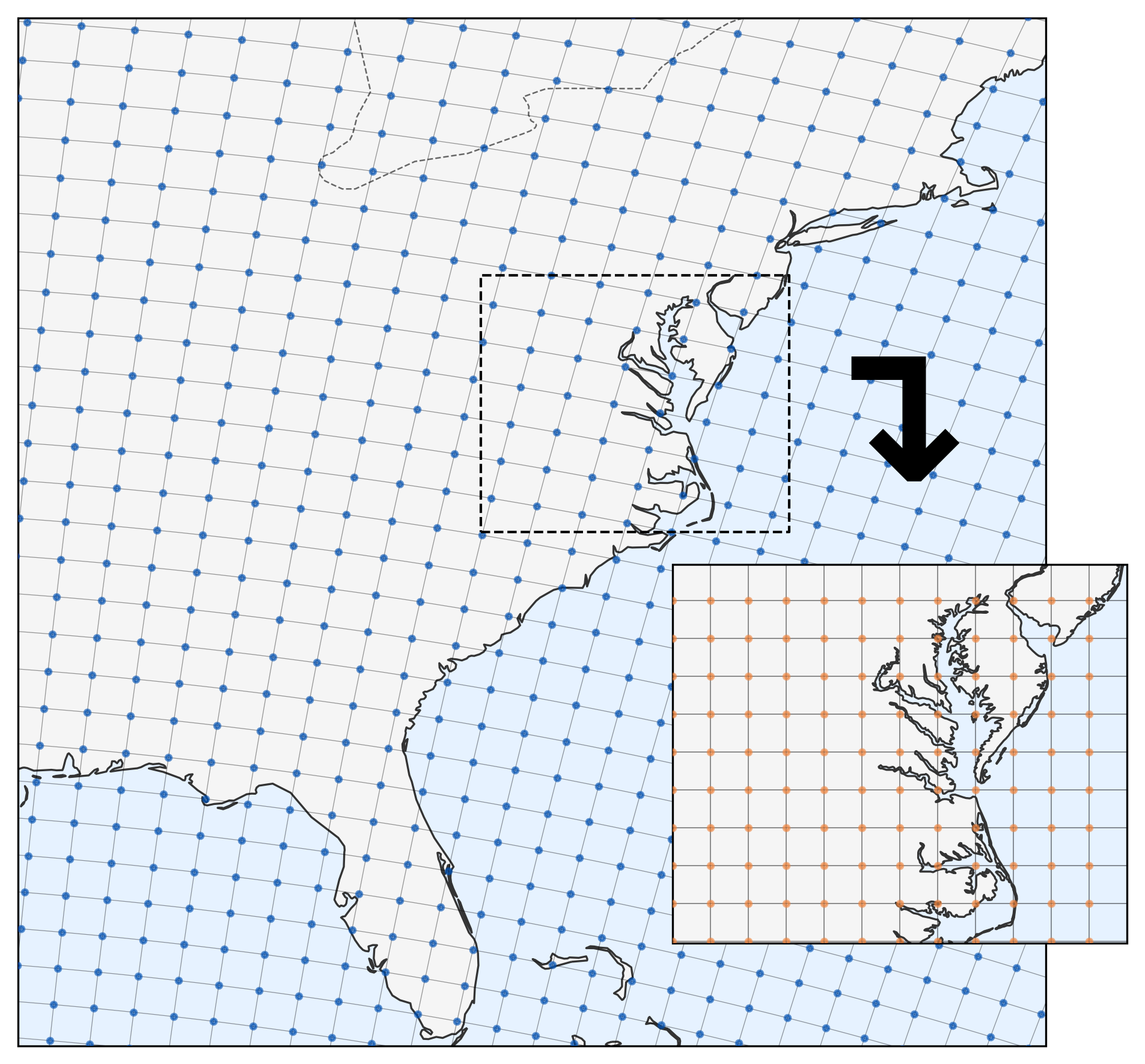}
    \caption{
    Illustration of the spatial transformation applied during regridding.
    }
    \label{fig:visual_transformation}
\end{figure}

Regridding must be physically aware: different atmospheric variables behave differently under spatial averaging or interpolation. For example, radiative fluxes and precipitation rates are extensive variables that depend on the underlying area -- their total value across a region must be conserved. In contrast, variables such as temperature or pressure are intensive, defined pointwise and independent of cell size. Wind-related variables pose an additional challenge due to the circular nature of direction and the vector character of speed components. To ensure that our downscaling model learns meaningful relationships rather than artifacts of inconsistent interpolation, the regridding strategy must respect these physical distinctions.
The resulting dataset consists of spatially aligned ERA5, WTK--US and HRRR for all selected variables. An example visualization is presented in (Figure~\ref{fig:visual_transformation}).

\section{Regridding - roundtrip error calculation} \label{appendix:Regridding-roundtrip-error}
The data is first interpolated from the native grid to the regular grid, and subsequently interpolated back to the original native grid using the corresponding reverse mapping.

This bidirectional interpolation framework enables isolation of the numerical error associated solely with the regridding procedure, avoiding discrepancies that would arise from comparisons between independent datasets. The resulting error represents the combined effects of information loss during the forward interpolation and additional distortions introduced during the backward interpolation. The reconstructed field is therefore directly compared to the original field on the native grid.

The round-trip interpolation error provides a conservative estimate of the regridding uncertainty, as it reflects the cumulative effects of both forward and backward mappings. In the training workflow, only the forward interpolation is applied; therefore, the effective error is expected to be smaller in practice.

Table~\ref{tab:rmse-mae-results} presents the roundtrip regridding error. The error remains low for temperature and wind variables but is substantially higher for surface pressure. Surface pressure is particularly sensitive to interpolation due to its strong dependence on elevation ($\approx -12$ hPa per 100 m), which is also reflected in the large standard deviation of this variable. Given the $\approx 2.2$ km grid spacing of our models, elevation changes within a single grid cell can already be significant, especially over complex terrain. Consequently, even small spatial misalignments introduced during interpolation may lead to noticeable pressure differences. These results suggest caution when comparing surface pressure errors across approaches evaluated on different grids or interpolation schemes. As part of future work, we plan to investigate more robust surface pressure regridding methods and strategies for reducing interpolation-induced errors.
\begin{table}[h]
\caption{
Summary of roundtrip regridding error across variables. Reported values are mean $\pm$ standard deviation. $T_2$ denotes 2-m air temperature ($^\circ$C), $u_{10}$ and $v_{10}$ wind component speeds (m/s), and $P$ surface pressure (hPa). \\
}
\label{tab:rmse-mae-results}
\centering
\setlength{\tabcolsep}{6pt}
\renewcommand{\arraystretch}{1.0}
\resizebox{\linewidth}{!}{
\begin{tabular}{@{}lcccc@{}}
\toprule
Metric & $T_2$ & $u_{10}$ & $v_{10}$ & $P $ \\
\midrule
RMSE & $0.248 \pm 0.051$ & $0.184 \pm 0.050$ & $0.182 \pm 0.053$ & $2.3439 \pm 40.386$  \\
MAE  & $0.117 \pm 0.024$ & $0.089 \pm 0.025$ & $0.090 \pm 0.028$& $0.9510 \pm 15.381$ \\
\bottomrule
\end{tabular}
}
\end{table}

\newpage
\section{HRRR Great Lakes Elevation Issue}\label{appendix:hrrr_great_lakes}

During exploratory analysis of the HRRR surface pressure field, we identified a temporally localized bias over the Great Lakes region. This artifact is consistent with the issue documented by the National Weather Service, which states that the December 2020 HRRR upgrade inadvertently specified the elevation of all Great Lakes grid points at 0m \cite{NOAA2022_hrrr_great_lakes}.  

\begin{figure}[b]
    \centering
    \includegraphics[width=\textwidth]{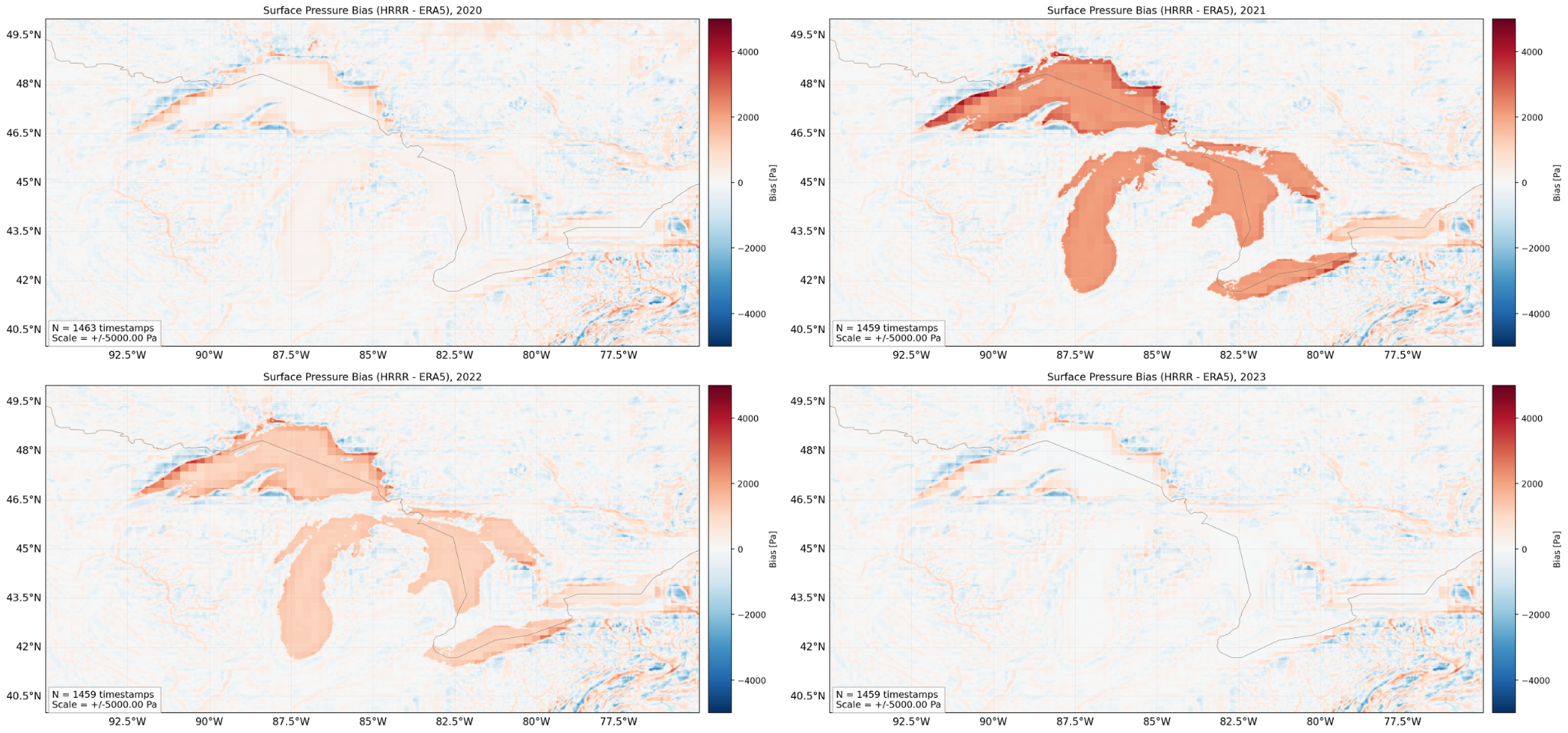}
    \caption{
Yearly mean surface pressure bias over the Great Lakes region, computed as HRRR minus ERA5, for 2020--2023. The positive bias is strongest in 2021, consistent with the documented HRRR Great Lakes elevation issue.}
    \label{fig:spatial_hrrr_great_lakes_pressure_bias}
\end{figure}

To quantify the impact of this issue, we compare HRRR surface pressure against ERA5 respective variable. Figure~\ref{fig:spatial_hrrr_great_lakes_pressure_bias} shows yearly mean maps of the surface pressure bias, computed as the difference between HRRR and ERA5, over the Great Lakes region. The bias pattern is weak in 2020, becomes clearly pronounced in 2021, remains visible in 2022, and is substantially reduced again in 2023. The spatial structure aligns closely with the Great Lakes boundaries, indicating that the anomaly is not a broad regional model bias but rather a localized artifact.

\begin{figure}[t]
    \centering
    \includegraphics[width=\textwidth]{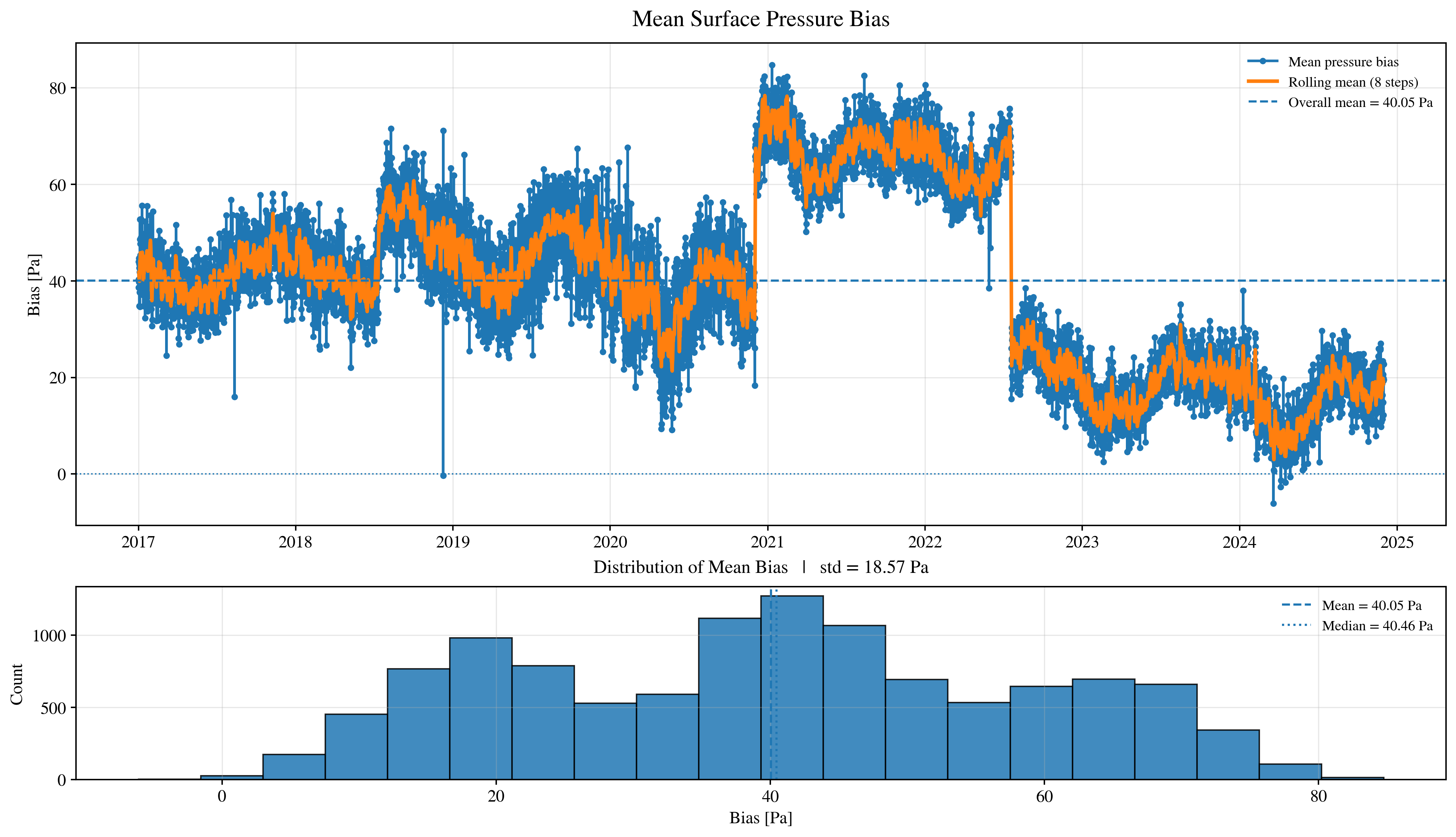}
    \caption{
    Full-domain time series and distribution of the spatially averaged HRRR \textemdash ERA5 surface pressure bias. The elevated bias during 2021--2022 is consistent with the documented Great Lakes elevation issue.
    }
    \label{fig:hrrr_great_lakes_pressure_bias}
\end{figure}

The temporal behavior is shown in Figure~\ref{fig:hrrr_great_lakes_pressure_bias}, where we plot the mean surface pressure bias averaged over the full HRRR domain. An abrupt increase occurs near the end of 2020, followed by a persistently elevated bias through 2021 and into 2022. After the correction was implemented in July 2022, the bias drops significantly and remains lower throughout 2023 and 2024. This discontinuity is consistent with the timing described in Service Change Notice and strongly suggests that the affected period should not be used for pressure-based model evaluation without additional filtering.

\begin{figure}[h]
    \centering
    \includegraphics[width=\textwidth]{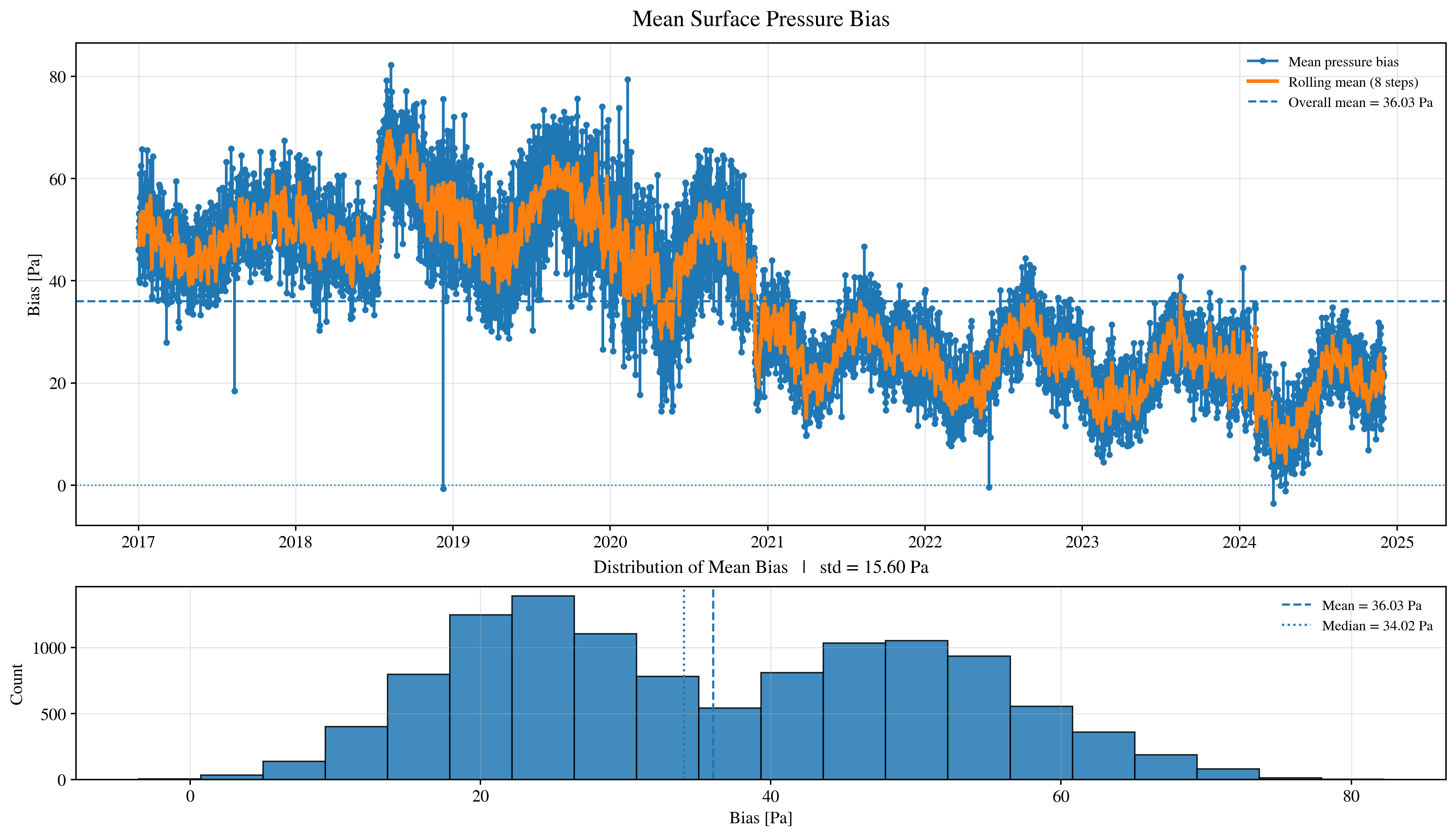}
    \caption{
    Same as Figure~\ref{fig:hrrr_great_lakes_pressure_bias}, but with the Great Lakes region excluded from the spatial average. The sharp shift seen in the full-domain time series is strongly reduced, demonstrating that the anomaly is primarily confined to the Great Lakes and is not representative of the broader HRRR domain.
    }
    \label{fig:hrrr_great_lakes_pressure_bias_region_excluded}
\end{figure}

To verify that this behavior is driven by the Great Lakes artifact rather than by a domain-wide change in HRRR quality, we repeat the same analysis after excluding the Great Lakes region. The resulting time series and distribution are shown in Figure~\ref{fig:hrrr_great_lakes_pressure_bias_region_excluded}. Once the Great Lakes are removed, the sharp discontinuity largely disappears and the mean bias becomes much more temporally consistent. This provides additional evidence that the anomaly is spatially confined and originates from the documented elevation error.

Based on these diagnostics, all surface-pressure experiments in this work exclude the affected interval from December 2020 through July 2022. This filtering avoids contaminating training and evaluation statistics with a known non-physical artifact and ensures that the reported pressure errors more accurately reflect model performance.

\newpage
\section{Inter-Dataset Bias Effects} \label{appendix:data-bias}

\begin{figure}[h]
    \centering
    \includegraphics[width=0.85\linewidth]{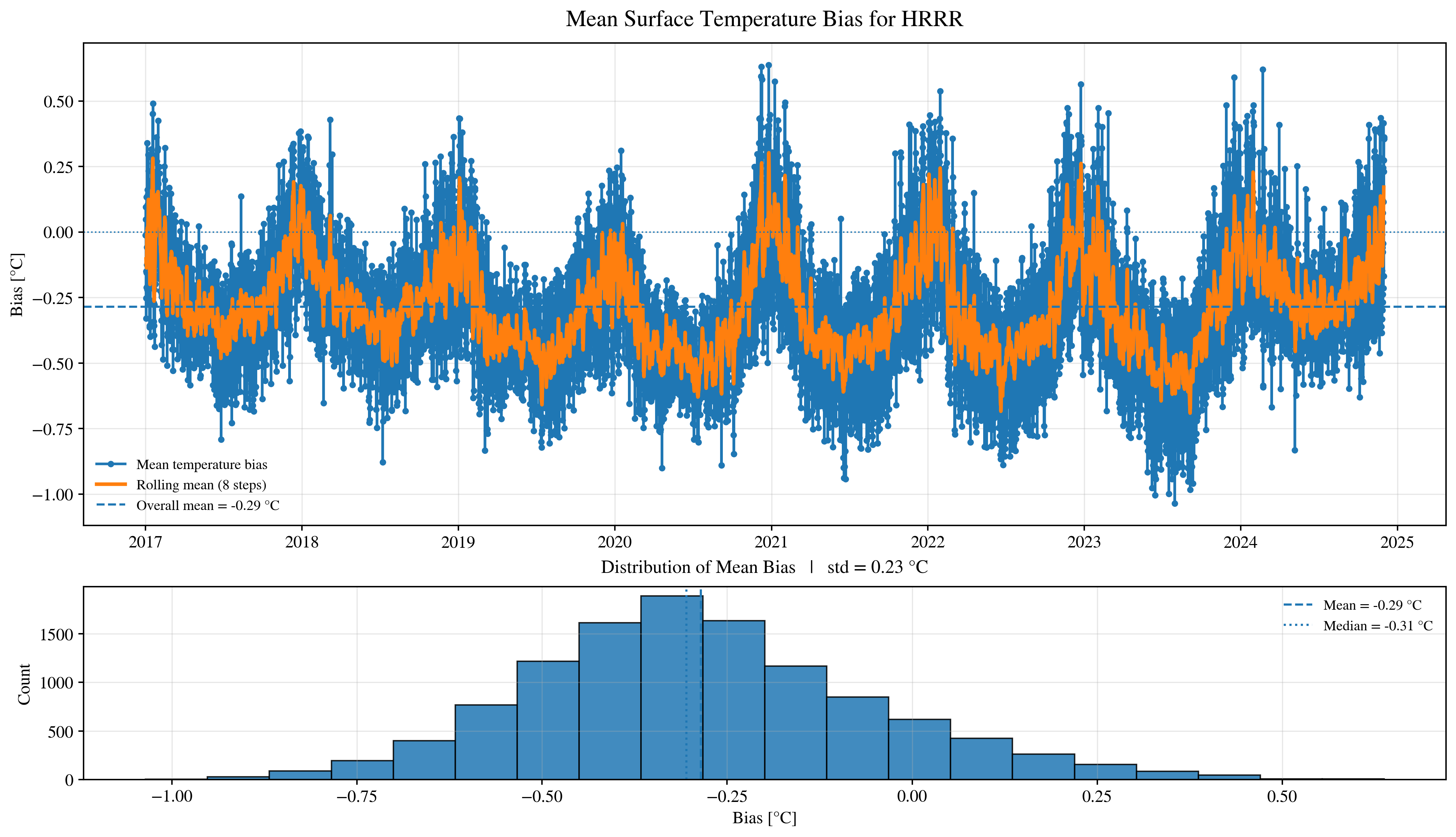}
    \caption{
    Temperature [$^\circ$C] bias off HRRR with respect to ERA5 dataset. Median for 8-years span is around -0.29 [$^\circ$C].
    }
    \label{fig:HRRR-bias}
\end{figure}

\begin{figure}[h]
    \centering
    \includegraphics[width=0.85\linewidth]{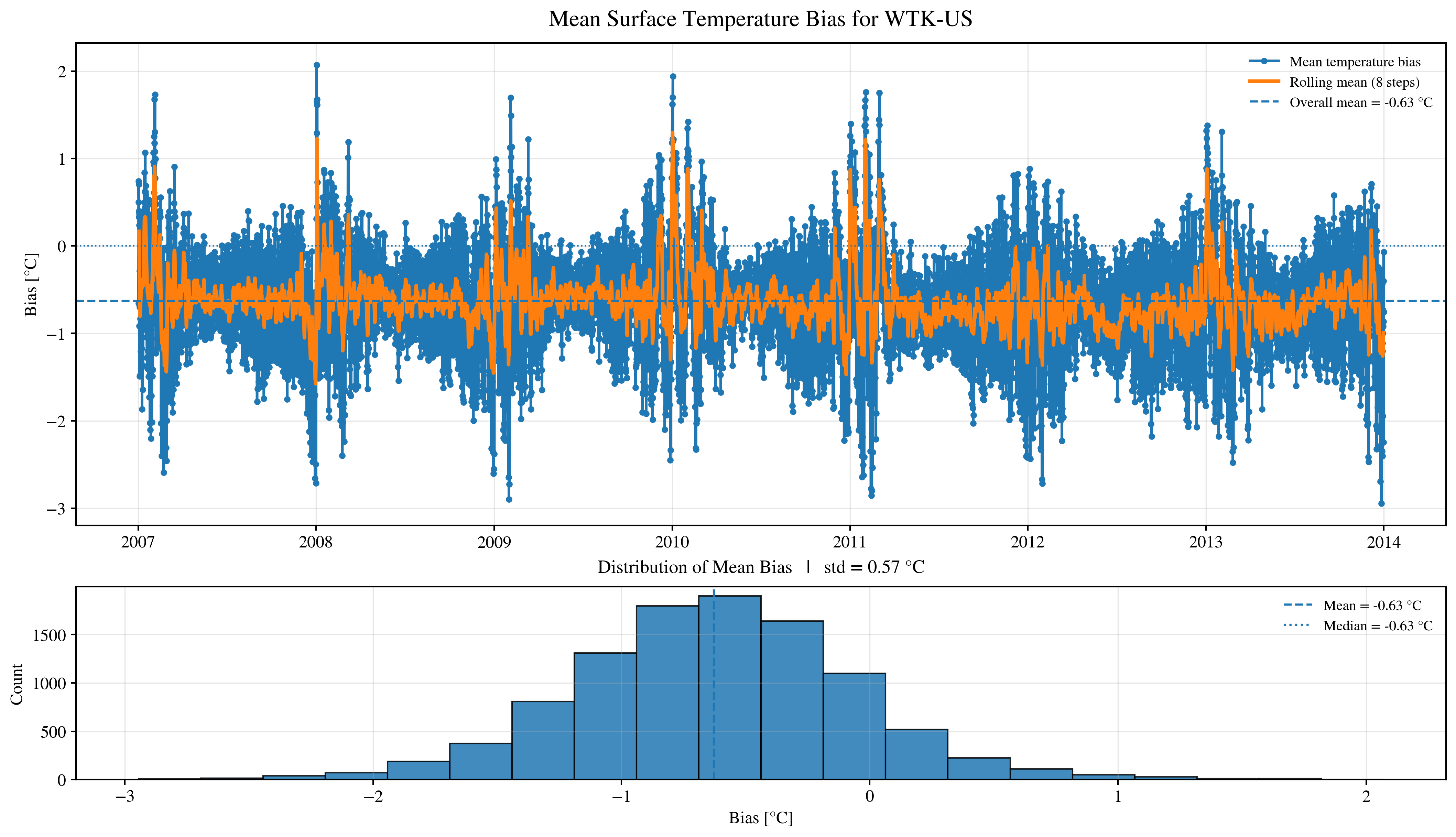}
    \caption{
    Temperature [$^\circ$C] bias off HRRR with respect to ERA5 dataset. Median for 7-years span is around -0.63 [$^\circ$C].
    }
    \label{fig:WTK-US-bias}
\end{figure}

In Figures~\ref{fig:HRRR-bias} and \ref{fig:WTK-US-bias}, we illustrate the respective dataset biases relative to ERA5. The WTK-US dataset exhibits a mean temperature bias of -0.63 [$^\circ$C], while HRRR shows a smaller bias of -0.29 [$^\circ$C]. Although the datasets do not overlap temporally, making direct comparison difficult, ERA5 serves here as a consistent reference due to its extensive validation and widespread use.

These differing biases affect the performance of our best model, $\mathrm{MLP_{LH}\text{-}40\text{-}2} \mid \mathrm{MLP\text{-}8\text{-}2\upscale{40}_{NN}}$, particularly when evaluated on the HRRR grid. During training, the model is exposed not only to HRRR but also to WTK-US, thereby inheriting bias from both datasets. This leads to degraded performance on HRRR-based evaluation.

\section{Embedding Correlation Analysis}\label{appendix:correlation-analysis}

\begin{figure*}[!h]

    \centering
    \includegraphics[width=\linewidth]{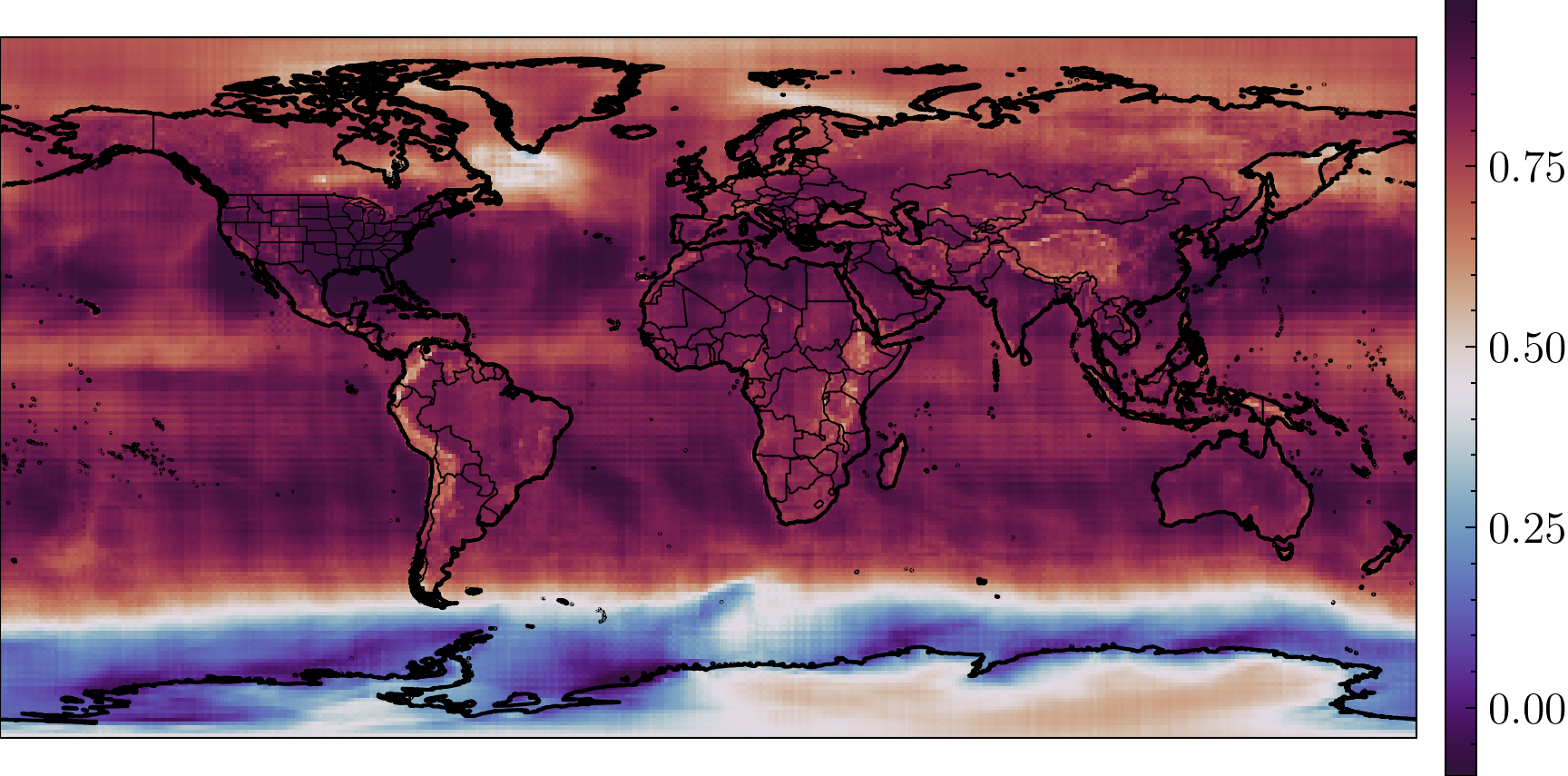}
    \caption{
        Embedding correlation plot. For each geographic location, we identify the embedding with the highest correlation within the CONUS region. This produces a visualization that maps, for every global position, the most similar embedding in the U.S., highlighting cross-regional embedding similarities.}
    \label{fig:distance compare}
\end{figure*}

\subsection{Embeddings similarity}

Each embedding is processed independently by Aurora's decoder, therefore, it must possess all required meteorological information within the given spatial patch and does not require data from other global regions. We anticipate high similarity among:
\begin{itemize}
    \item Embeddings spatially proximate to each other.
    \item Embeddings corresponding to identical seasonal periods.
    \item Embeddings derived from analogous geographical regions (e.g., oceanic areas exhibit higher mutual similarity than flatlands).
\end{itemize}

To enable cross-regional comparison while accounting for temporal offsets from time zone variations, embeddings from identical timestamps cannot be compared directly with each other. For analysis, we first average every 28 consecutive embeddings (representing one week) before computing similarity metrics.

\paragraph{Distance Similarity}
Figure~\ref{fig:distance compare} shows that the correlations between global climate and the \textit{CONUS} region occur at \textit{CONUS} latitudes ($-25 ^\circ\,\mathrm{to}\,-45 ^\circ $ in the northern hemisphere and $25 ^\circ \,\mathrm{to}\,45 ^\circ $ in the southern hemisphere). The weakest correlations appear in polar regions and equatorial oceans.
These results indicate that Aurora's backbone embeddings preserve the expected distance similarity characteristics.

\paragraph{Temporal Similarity}

\begin{figure}[!t]
    \centering
    \begin{subfigure}{.5\textwidth}
      \centering
      \includegraphics[width=.9\linewidth]{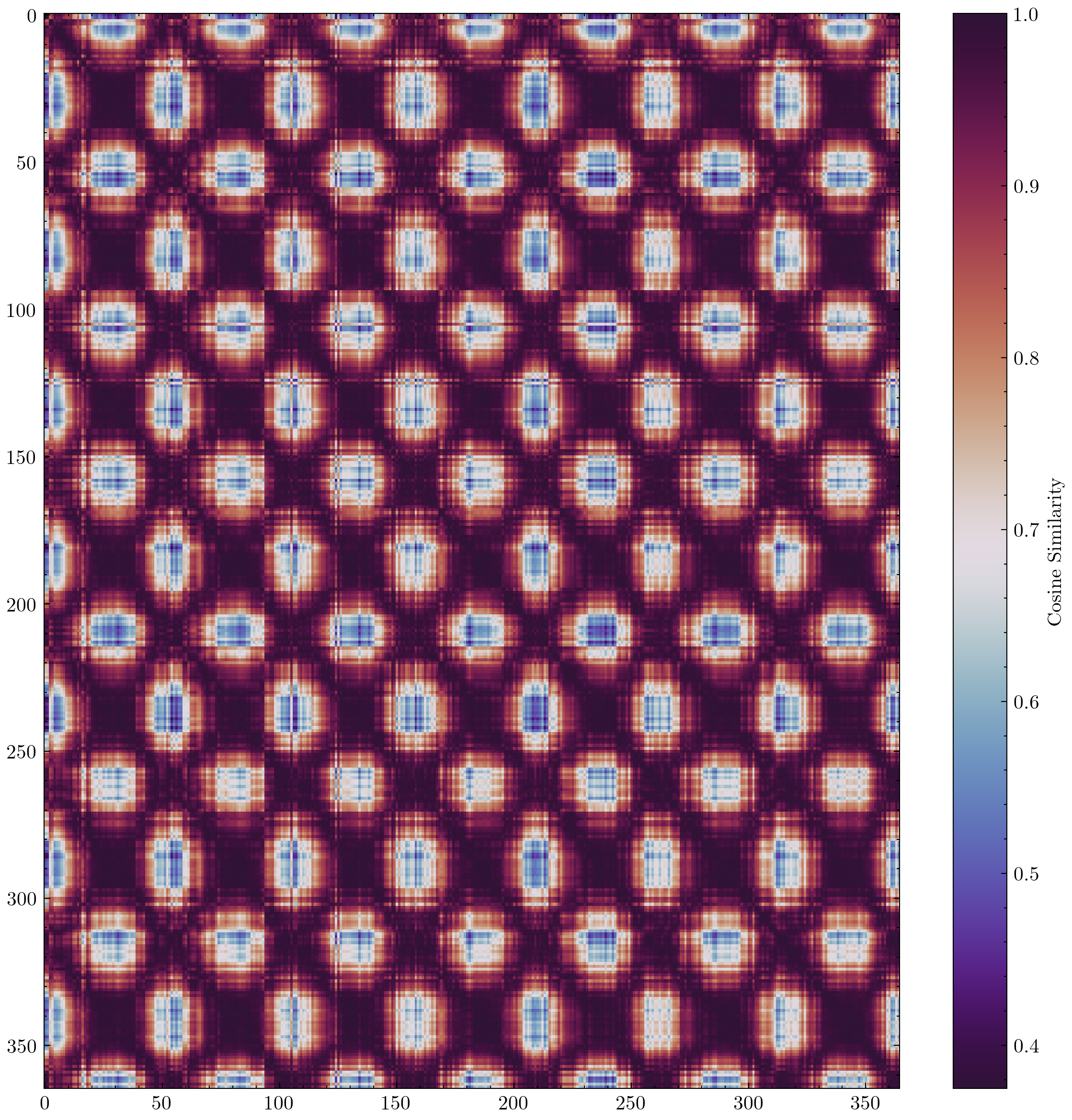}
      \caption{Surface variables temporal correlation}
      \label{fig:sfig1}
    \end{subfigure}%
    \begin{subfigure}{.5\textwidth}
      \centering
      \includegraphics[width=.9\linewidth]{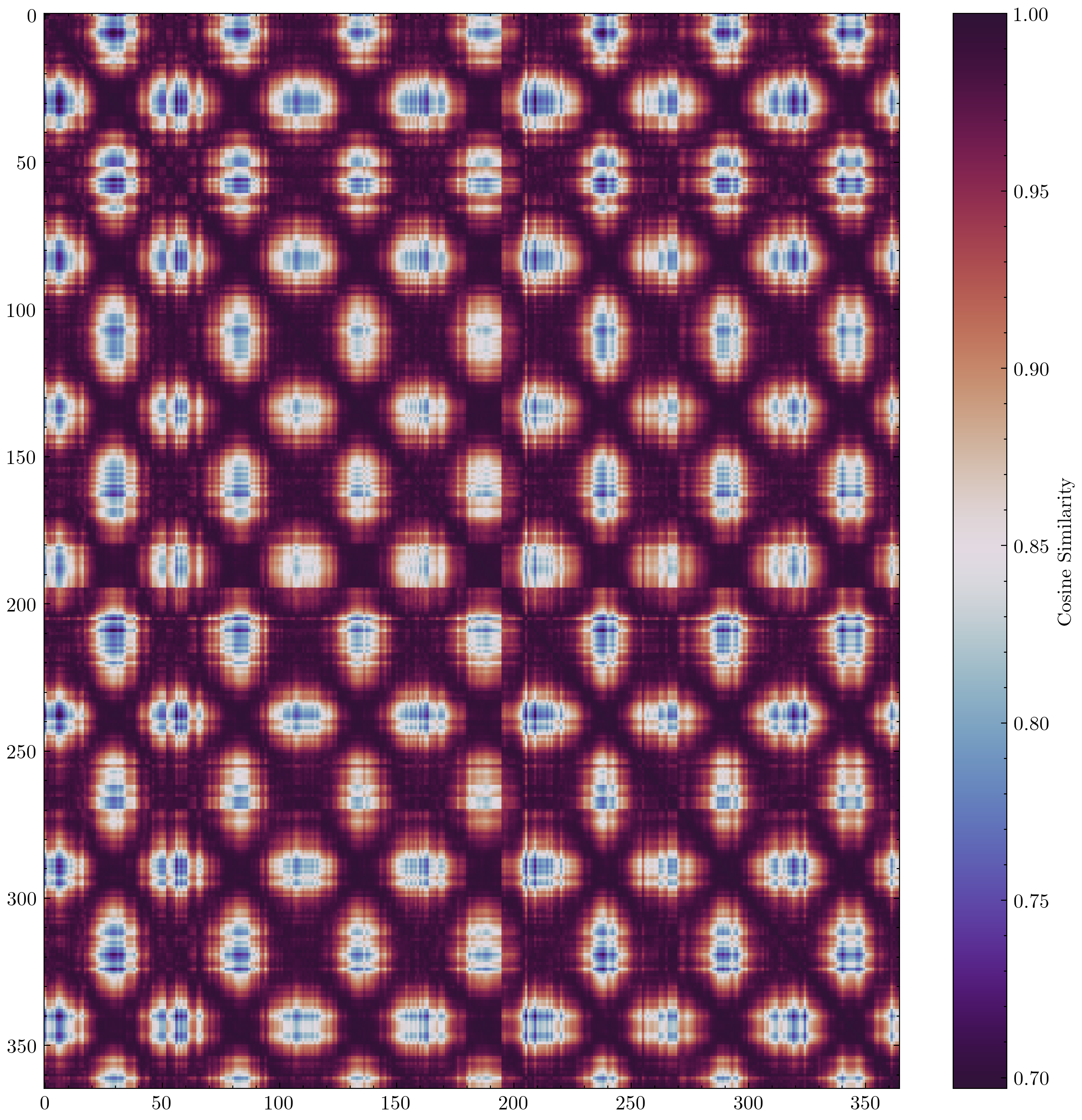}
      \caption{Atmospheric variables temporal correlation}
      \label{fig:sfig2}
    \end{subfigure}
    \caption{
    Comparison of temporal correlation between surface and atmospheric channels of Aurora's embeddings, spatially averaged over the \textit{CONUS} region.
    }
    \label{fig:temporal-similarity}
\end{figure}

We evaluate temporal similarity of the embeddings by first spatially averaging across the entire CONUS region and temporally averaging every $28$ consecutive embeddings (corresponding to one week). Pairwise similarity is then computed between these weekly averaged embeddings over the 2007--2013 period. Figure~\ref{fig:temporal-similarity} demonstrates distinct separation between summer and winter patterns, along with interannual seasonal consistency, for both surface-level and atmospheric variables.

Notably, the similarity plots exhibit irregularly spaced lines characterized by markedly reduced similarity values. This phenomenon warranted further investigation, as it suggested the presence of a potential systematic effect, either originating from the input data or from the Aurora model itself.

The following observations were established:
\begin{itemize}
    \item The low-similarity lines occur on few different dates and typically persist for groups of 2–4 consecutive timesteps.
    \item No consistent relationship was identified between the season and the occurrence of the phenomenon.
    \item Although some occurrences coincide with periods of snowfall or heavy precipitation, no causal relationship between these meteorological conditions and the low-similarity events was observed.
\end{itemize}
These findings do not indicate any meaningful seasonal or meteorological dependence. Consequently, we examined potential dependencies related to the data acquisition process. The following observations were made:
\begin{itemize}
    \item The satellite \textit{Metop-A} became operational on 21 May 2007, as reported by ECMWF~\cite{schluessel2007metop}. This date coincides with the first observed occurrence of the low-similarity phenomenon.
    \item \textit{Metop-A} contributes observations to the ERA5 dataset, on which Aurora was trained, as documented by ECMWF~\cite{era5_observations}.
    \item The final low-similarity line visible in the similarity plot coincides with the operational handover of \textit{Metop-B}, the direct successor of \textit{Metop-A}~\cite{esa_metop_handover2012}.  
\end{itemize}
Altogether, these observations suggest that the introduction of new satellite sensors may have introduced subtle artifacts into the input data. While such artifacts may be visually insignificant for human interpretation, they can be consequential for machine learning models operating on high-dimensional representations. We therefore hypothesize that calibration inconsistencies during the initial operational phases of newly deployed sensors may have temporarily affected the data distribution, thereby influencing the embedding generation process.

\begin{figure}
    \centering
    \includegraphics[width=0.9\linewidth]{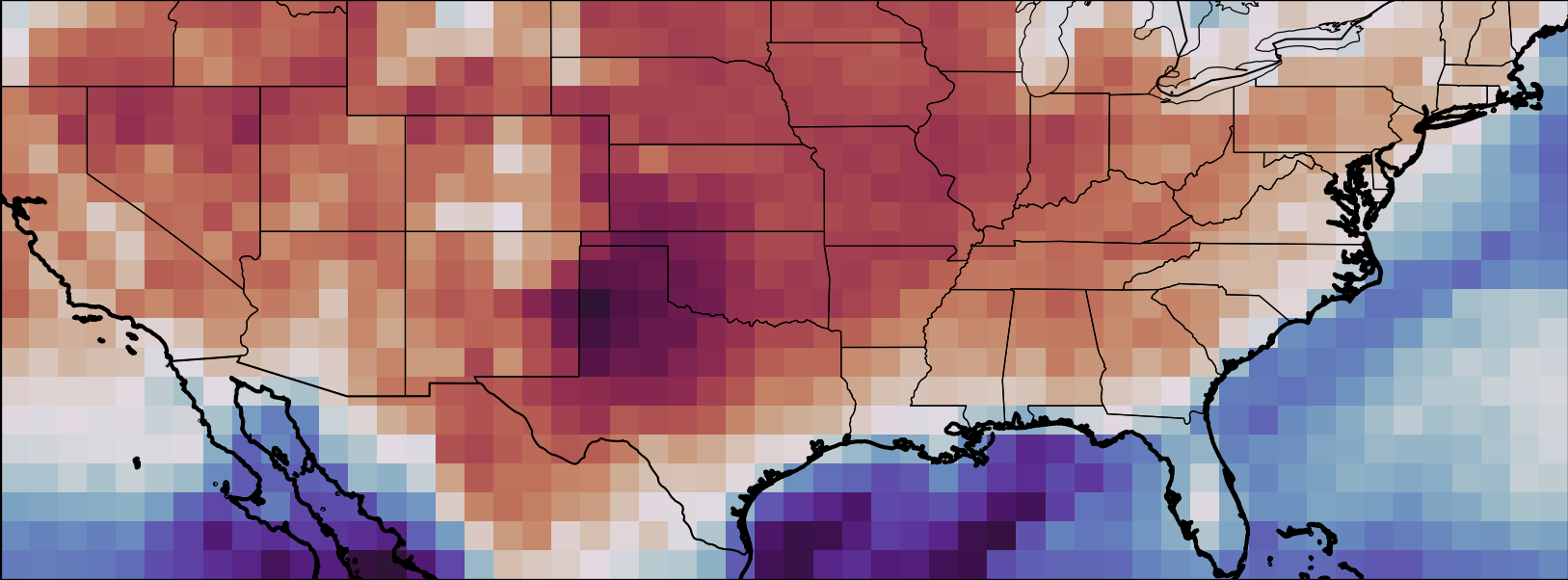}
    \caption{Pairwise cosine similarity between a reference patch embedding and embeddings of all other patches.}
    \label{fig:area-similarity}
\end{figure}

\paragraph{Terrain-Dependent Spatial Similarity in the Latent Space}
As discussed above, embeddings are expected to exhibit similarity that depends not only on spatial proximity but also on terrain characteristics (e.g., mountains, oceans, flatlands). To investigate this hypothesis, we randomly select a single patch within the CONUS region and compare its cosine similarity to all other patches in the region, averaged over the first week of May 2010. This temporal aggregation mitigates short-term meteorological variability, the specific time period is otherwise not critical for the experiment.
The resulting similarity map is shown in Figure~\ref{fig:area-similarity}. As expected, the central patch exhibits the highest (self-)similarity. Moving away from this location, similarity generally decreases, however, the decay is markedly non-isotropic. This anisotropic pattern indicates that the relationship between spatial distance and embedding similarity is non-linear and strongly modulated by underlying terrain. In particular, sharp similarity gradients are observed along coastlines and over oceanic regions, where correlations attain their lowest values. These findings suggest that terrain topography plays a substantial role in shaping the structure of the latent space beyond mere geographic distance.

\section{Energy Spectra Analysis}\label{appendix:spectra}
\begin{figure*}[!h]

    \centering
    \includegraphics[width=\linewidth]{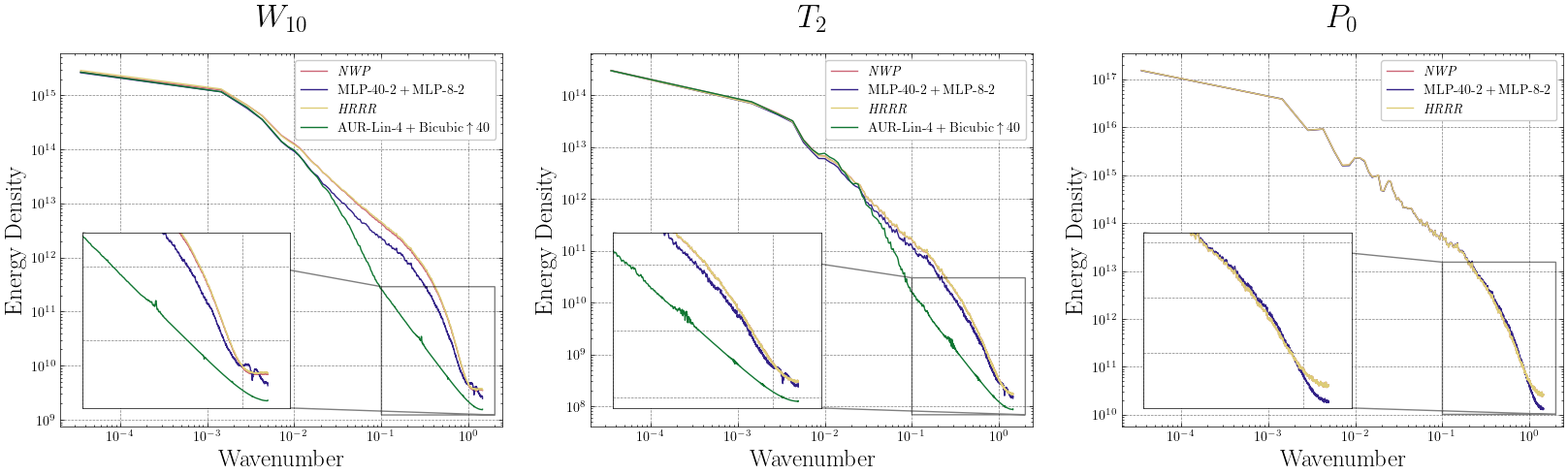}
    \caption{
        Energy spectra of surface variables computed on the HRRR grid and averaged over the evaluation set. The wavenumber is expressed in units of inverse grid points, such that $10^{-1}$ corresponds to a wavelength $\lambda = 21.75\,\mathrm{km}$.}
    \label{fig:energy-spectra}
\end{figure*}

Figure \ref{fig:energy-spectra} shows energy spectra averaged over the evaluation year 2022 for each variable we aim to predict. Shown are: NWP forecasts 6h ahead computed with the WRF-ARW model; our best model, $\mathrm{MLP_{LH}\text{-}40\text{-}2} \mid \mathrm{MLP\text{-}8\text{-}2\upscale{40}_{NN}}$; and Aurora base model outputs combined with standard bicubic upscaling to match the resolution of the other methods. All are compared against ground truth data from HRRR. Both our model and the NWP forecasts closely follow the ground truth, particularly for pressure and temperature variables. This indicates that decoders do not loose physical meaning introduced by Aurora's backbone and are somewhat accurate in representing the underlying physics. As expected, $\mathrm{AUR\text{-}MLP\text{-}4\text{-}0\upscale{40}_{Bicubic}}$, due to its coarser grid resolution, diverges around $\mathrm{10^{-3/2}}$, reflecting model dissipation beyond the scale at which it can reliably represent the system. The tails observed, particularly in the pressure spectrum, reflect a buildup of energy to physically unrealistic levels, an effect expected in both the ML approaches and mesoscale NWP simulations at the highest wavenumbers.

\newpage
\section{Distribution of error on in-situ observations} \label{appendix:in-situ}

Evaluating against in situ observations from \textit{HadISD}, we find that our model consistently outperforms \textit{Aurora} across the \textit{CONUS} region. This improvement is particularly pronounced in areas that are difficult to model, such as mountainous terrain with sharply varying elevations and regions influenced by strong land–water contrasts (e.g., along the California coast and around the Great Lakes), where associated mesoscale processes render weather forecasting more challenging. In these regions, atmospheric conditions exhibit strong spatial variability; therefore, fine-resolution modelling is crucial to adequately capture physical processes that are not resolved at the coarser $0.25^\circ$ scale. This observation is consistent for both temperature and wind.

\begin{figure*}[!h]
    \centering
    \begin{minipage}[t]{0.48\linewidth}
        \centering
        \includegraphics[width=\linewidth]{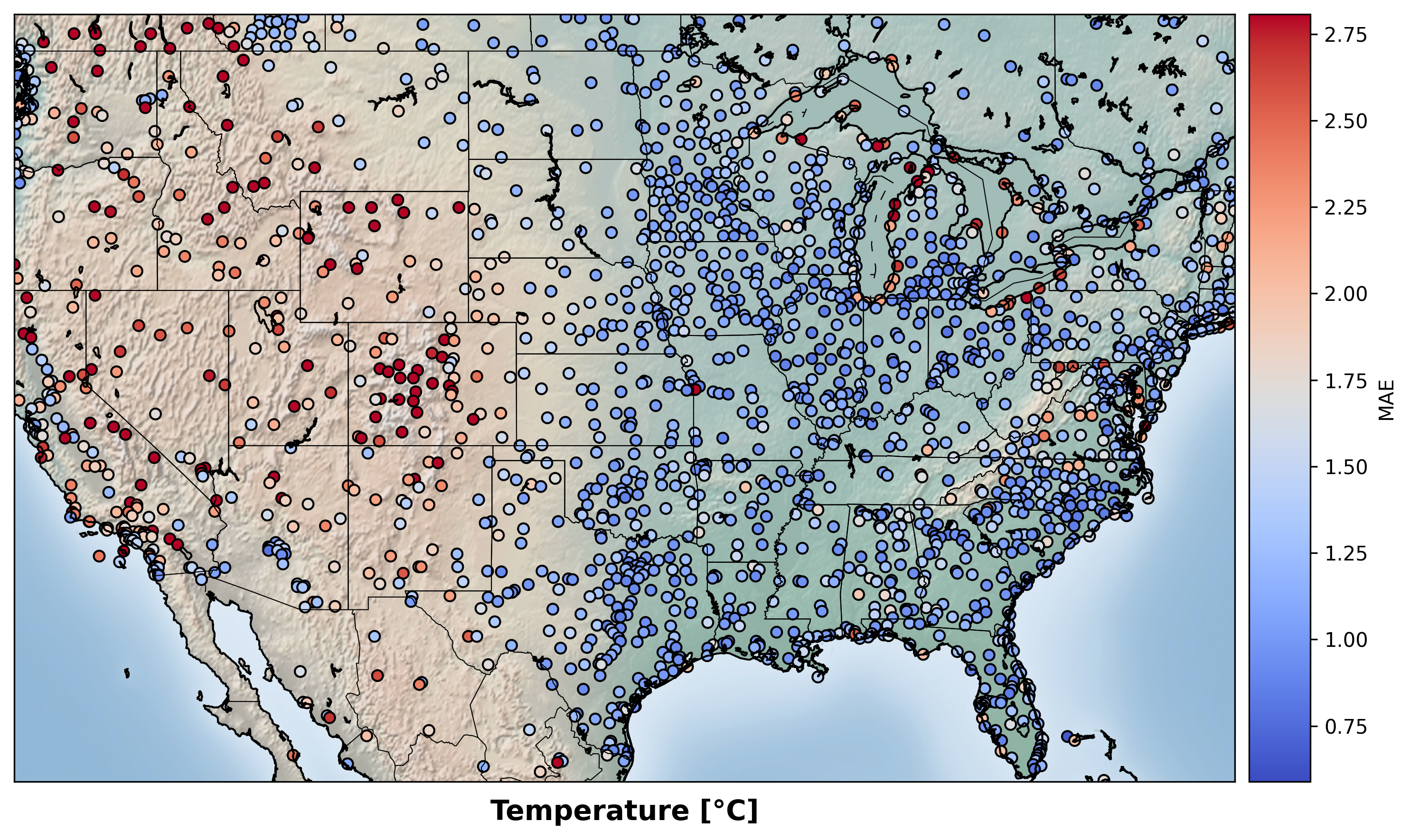}
    \end{minipage}
    \hfill
    \begin{minipage}[t]{0.48\linewidth}
        \centering
        \includegraphics[width=\linewidth]{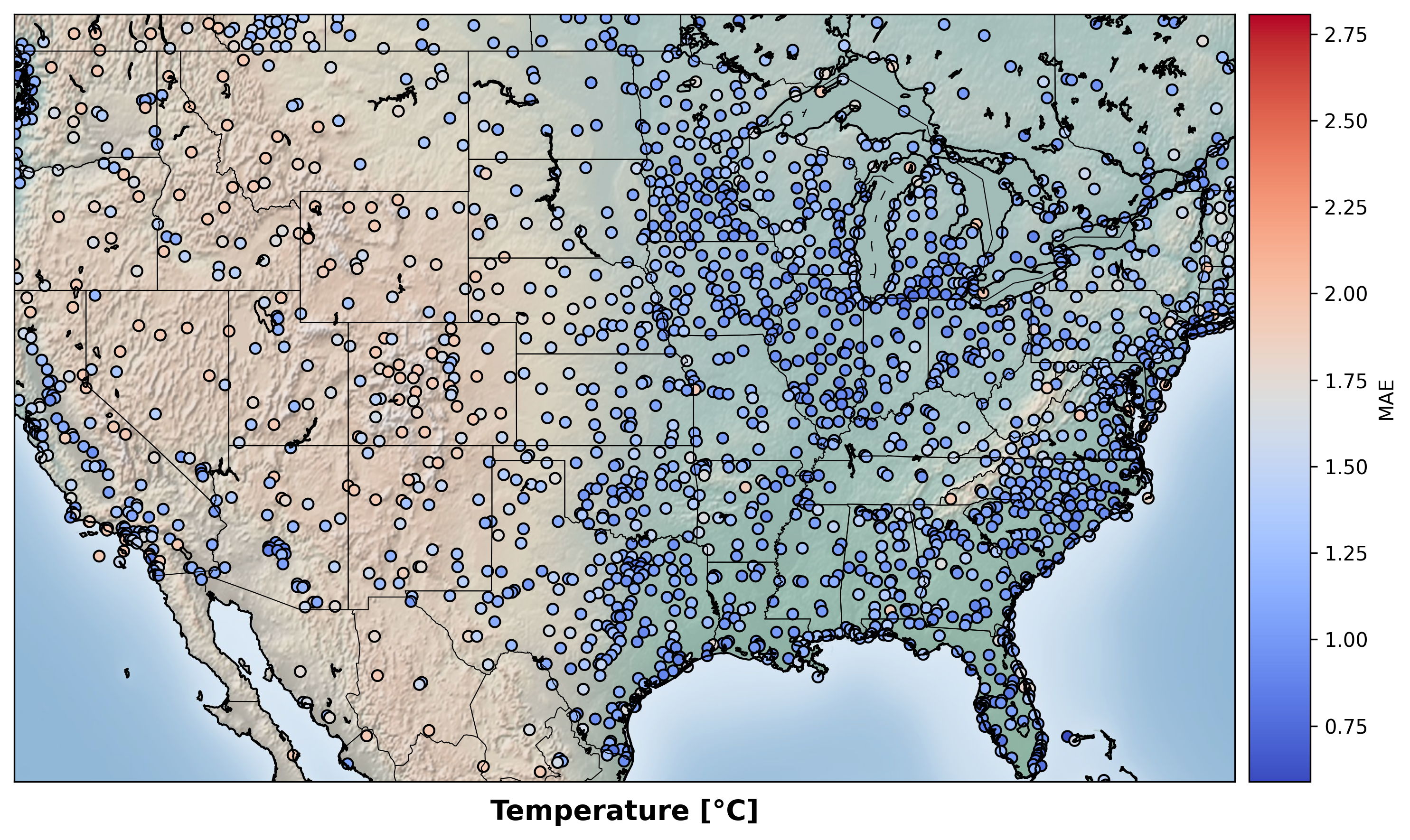}
    \end{minipage}
    \caption{
        Distribution of temperature \textit{MAE} between \textit{HadISD} weather station observations and model predictions for the 2022 evaluation year: Aurora predictions (left) and our best model (right).
    }
    \label{fig:temperature_2m_distribution}
\end{figure*}

\begin{figure*}[!h]
    \centering
    \begin{minipage}[t]{0.48\linewidth}
        \centering
        \includegraphics[width=\linewidth]{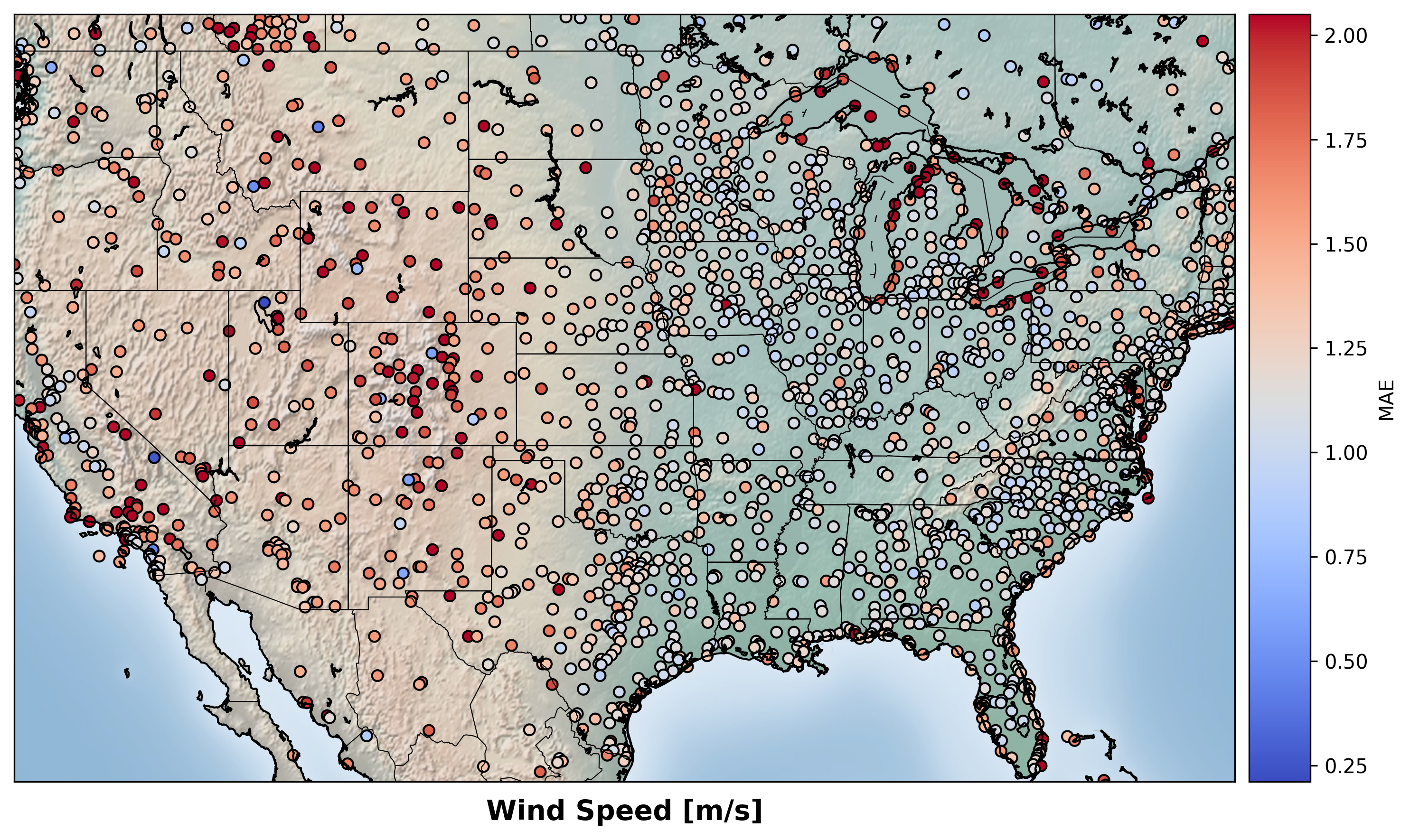}
    \end{minipage}
    \hfill
    \begin{minipage}[t]{0.48\linewidth}
        \centering
        \includegraphics[width=\linewidth]{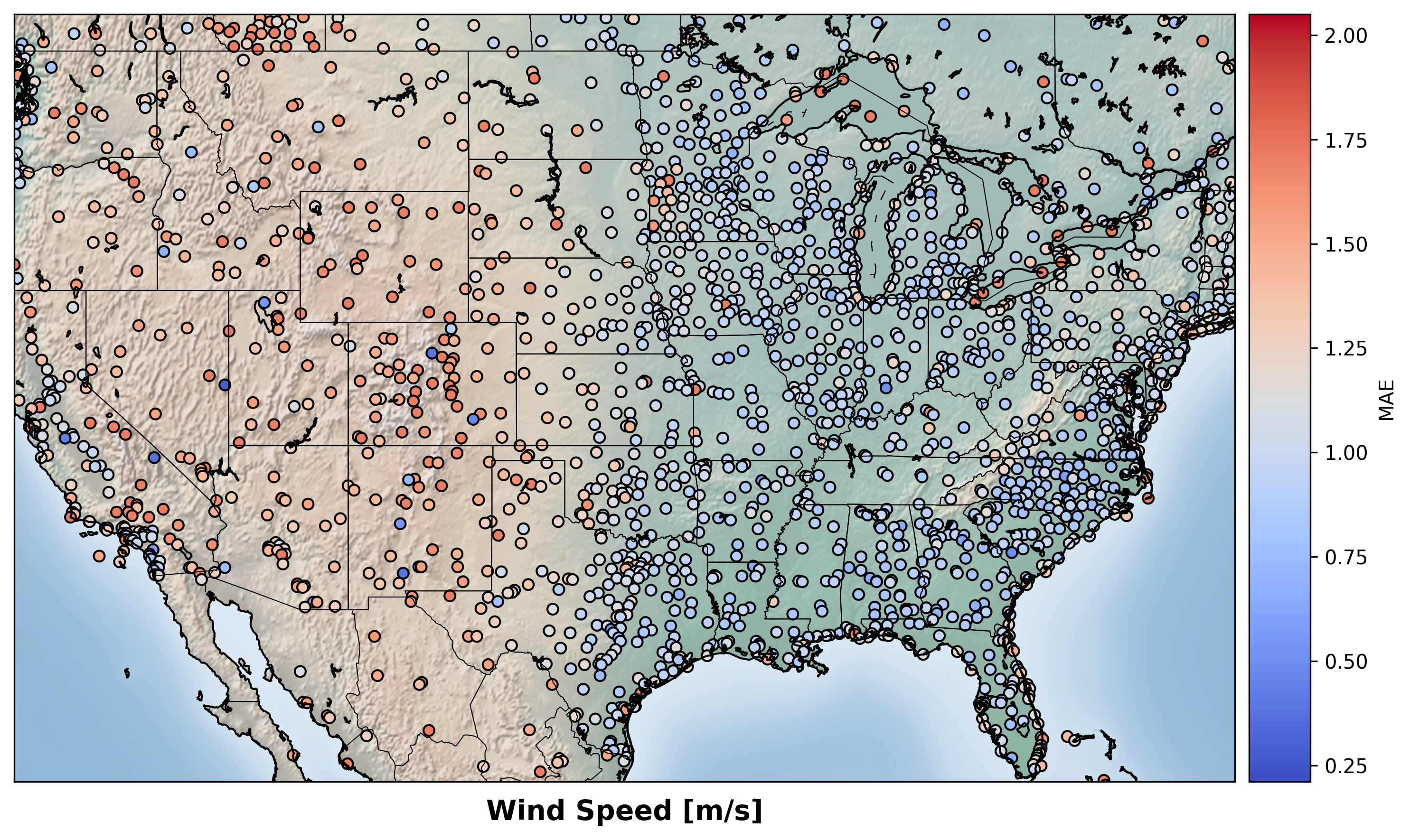}
    \end{minipage}
    \caption{ 
    Distribution of wind \textit{MAE} between \textit{HadISD} weather station observations and model predictions for the 2022 evaluation year: Aurora predictions (left) and our best model (right).
    }
    \label{fig:wind_distribution}
\end{figure*}

Figures~\ref{fig:temperature_2m_distribution} and \ref{fig:wind_distribution}, which illustrate these results, use a shared colour scale for each variable to facilitate comparison between models. For each variable, the colour scale spans from the minimum \textit{MAE} observed across both models to the larger of their 95th-percentile \textit{MAE} values. Values above the 95th percentile are capped for visualization purposes. A background elevation map is overlaid on the visualizations to highlight error patterns associated with mountainous terrain.

\newpage
\section{Additional results}
\paragraph{Multi-region training} \label{appendix:multiregional}

To measure our model's ability to adapt to multiple domains via cross-regional training we introduce a second geographically separate dataset DANRA \cite{Danra}. DANRA is a reanalysis dataset for Denmark and the surrounding Baltic and North Sea regions, which we will refer to as the Baltic region.
Similarly to the \textit{WTK-US} dataset it was generated by an \textit{NWP} model and is defined on LCC projected grid with spatial resolution of 2.5km and temporal resolution of 3h.

As a benchmark of how good prediction in the Baltic region can be, we train two types of models:

\begin{itemize}
    \item Trained solely on \textit{DANRA} dataset.
    \item Trained on both \textit{DANRA} and \textit{CONUS} datasets.
\end{itemize}

The first of the experiments is meant to provide the best possible result that can be achieved by our models for the Baltic region. The second one should show the potential of training a model on two different regions of the world simultaneously. \textit{CONUS} and \textit{DANRA} are quite representative examples of regions that models might be trained on, as they are far away from each other. 

Multi-region training results are summarized in Table~\ref{tab:multi-region}, which shows the performance when training exclusively on one region versus training on the combined dataset. We observe that a unified model trained on both CONUS and Danish regions successfully adapts to both domains. Performance remains comparable with only marginal increases in error. Notably, for the DANRA region, the combined training yields consistent improvements across $u$ and $v$ wind components. 

This demonstrates that our lightweight decoder heads capable of generating accurate weather predictions for two different regions located over 7000 kilometres away from each other, on two different continents and two different hemispheres. The large spatial separation between these regions reflects fundamentally different underlying physical processes governing their weather dynamics, thereby making the achieved performance particularly meaningful.

\begin{table}[]
\caption{Multi-region joint training across two geographically distinct regions. We evaluate the effect of training on both CONUS and another region, Denmark. Similarly to \ref{tab:wrf-arw}, we report RMSE for $T_2$, $u_{10}$, and $v_{10}$ wind components (m/s), as well as surface pressure $P$ (hPa). Additionally, we report the difference in RMSE relative to single-region training.
\\}

\label{tab:multi-region}
\centering
\small
\setlength{\tabcolsep}{3pt}
\renewcommand{\arraystretch}{1.1}
\begin{tabular}{@{}lrrrrrrrr@{}}
\toprule
& \multicolumn{4}{c}{CONUS region} & \multicolumn{4}{c}{Danish region} \\
\cmidrule(lr){2-5} \cmidrule(lr){6-9}
Training region & $T_2$  & $u_{10}$ & $v_{10}$  & $P$ & $T_2$  & $u_{10} $ & $v_{10} $ & $P $ \\
\midrule
CONUS & 0.8079 & 0.8819 & 0.9210 & 0.3494 & -- & -- & -- & -- \\
Danish & -- & -- & -- & -- & 0.5137 & 0.7364 & 0.7484 & 0.2227 \\
\makecell[l]{CONUS + Danish \\ Difference} & \makecell[r]{0.8249\\ +0.0170} & \makecell[r]{0.8916 \\ +0.0097} & \makecell[r]{0.9262 \\ +0.0052} & \makecell[r]{0.3841 \\ +0.0347} & \makecell[r]{0.5311\\+0.0174} & \makecell[r]{0.6708 \\ -0.0656} & \makecell[r]{0.6790 \\ -0.0694} & \makecell[r]{0.2376 \\ +0.0148} \\
\bottomrule
\end{tabular}
\footnotesize
\parbox{\linewidth}{
\raggedright
}
\end{table}

\paragraph{Evaluation on StormCast grid} \label{appendix:StormCast}
Due to the substantial roundtrip regridding error (see Appendix~\ref{appendix:Regridding-roundtrip-error}) relative to the performance gap between our model, $\mathrm{MLP_{LH}\text{-}40\text{-}2} \mid \mathrm{MLP\text{-}8\text{-}2\upscale{40}_{NN}}$, and the StormCast 5-member Ensemble, presented in Table \ref{tab:wrf-arw} we provide additional evaluation results in Table~\ref{tab:wrf-arw-othergrid}. The table reports two sets of results: one on our latitude–longitude grid and another on the LCC grid, where our predictions are regridded onto the StormCast grid. Although this introduces additional interpolation error, it enables a fairer comparison between the approaches and indicates that the regridding error is not large enough to materially affect the overall conclusions. Pressure results are not reported, as the StormCast was trained on the 2020 dataset.

\begin{table}[t]
\caption{
Performance evaluation of our approach against WRF-ARW, StormCast and Aurora’s $0.25^\circ$ base model. Results are presented for HadISD in-situ observations and the HRRR grid using RMSE at +6h lead time. 
$T_2$ denotes 2-m air temperature ($^\circ$C), $u_{10}$ and $v_{10}$ wind component speeds (m/s).
\\}
\label{tab:wrf-arw-othergrid}
\centering
\small
\setlength{\tabcolsep}{4pt}
\renewcommand{\arraystretch}{1.1}
\resizebox{0.95\linewidth}{!}{%
\begin{tabular}{@{}llrrrrrr@{}}
\toprule
Grid & Model & \multicolumn{3}{c}{HadISD (in-situ)} & \multicolumn{3}{c}{HRRR Grid} \\
\cmidrule(lr){3-5} \cmidrule(lr){6-8}
& & $T_2$  & $u_{10}$ & $v_{10}$  & $T_2$  & $u_{10} $ & $v_{10} $ \\
\midrule
& $\mathrm{AUR\text{-}MLP\text{-}4\text{-}0\upscale{40}_{NN}}$ &  &  &  &  &  &  \\
LCC (km) & \hspace*{0.4cm} $\mid \mathrm{MLP\text{-}40\text{-}2} \mid \mathrm{MLP\text{-}8\text{-}2\upscale{40}_{NN}}$ & 1.718 & 1.530 & 1.606 & 1.397 & 1.249 & 1.309 \\
Lat-lon ($^\circ$) & \hspace*{0.4cm} $\mid \mathrm{MLP\text{-}40\text{-}2} \mid \mathrm{MLP\text{-}8\text{-}2\upscale{40}_{NN}}$ & 1.708 & 1.540 & 1.615 & 1.430 & 1.262 & 1.346 \\
LCC (km) & StormCast & 2.538 & 1.972 & 2.063 & 2.352 & 1.757 & 1.836 \\
Lat-lon ($^\circ$) & StormCast  & 2.530 & 1.955 & 2.054 & 2.371 & 1.738 & 1.842 \\
\bottomrule
\end{tabular}
}
\end{table}

\newpage

\section{Data Scale and Computational Setup}\label{appendix:data-compute-setup}

\paragraph{Data volume and preprocessing.}
The complete training corpus comprises approximately 30~TB of raw meteorological fields. Directly operating on this data during training would incur substantial I/O and memory overhead. To address this, we precompute and cache intermediate embeddings produced by Aurora's frozen backbone from ERA5 inputs. These cached embeddings occupy approximately 350~GB of storage and are reused across all experiments, enabling efficient training of multiple decoder configurations without repeated backbone inference.

\paragraph{Computational resources.}
All experiments were conducted on NVIDIA H100 GPUs. A single training run requires approximately 10 hours, while inference on the same hardware takes around 1 minutes. By training exclusively on cached Aurora embeddings rather than raw ERA5 fields, GPU memory usage remains modest at approximately 20~GB. The dominant memory requirement arises from storing cached embeddings in system memory, with a total RAM footprint of approximately 350~GB. All computations are performed in single-precision (FP32).

\section{Performance vs. Training Data Size} \label{appendix:scaling-laws}
\begin{figure}[!t]
    \centering
    \includegraphics[width=1.0\linewidth]{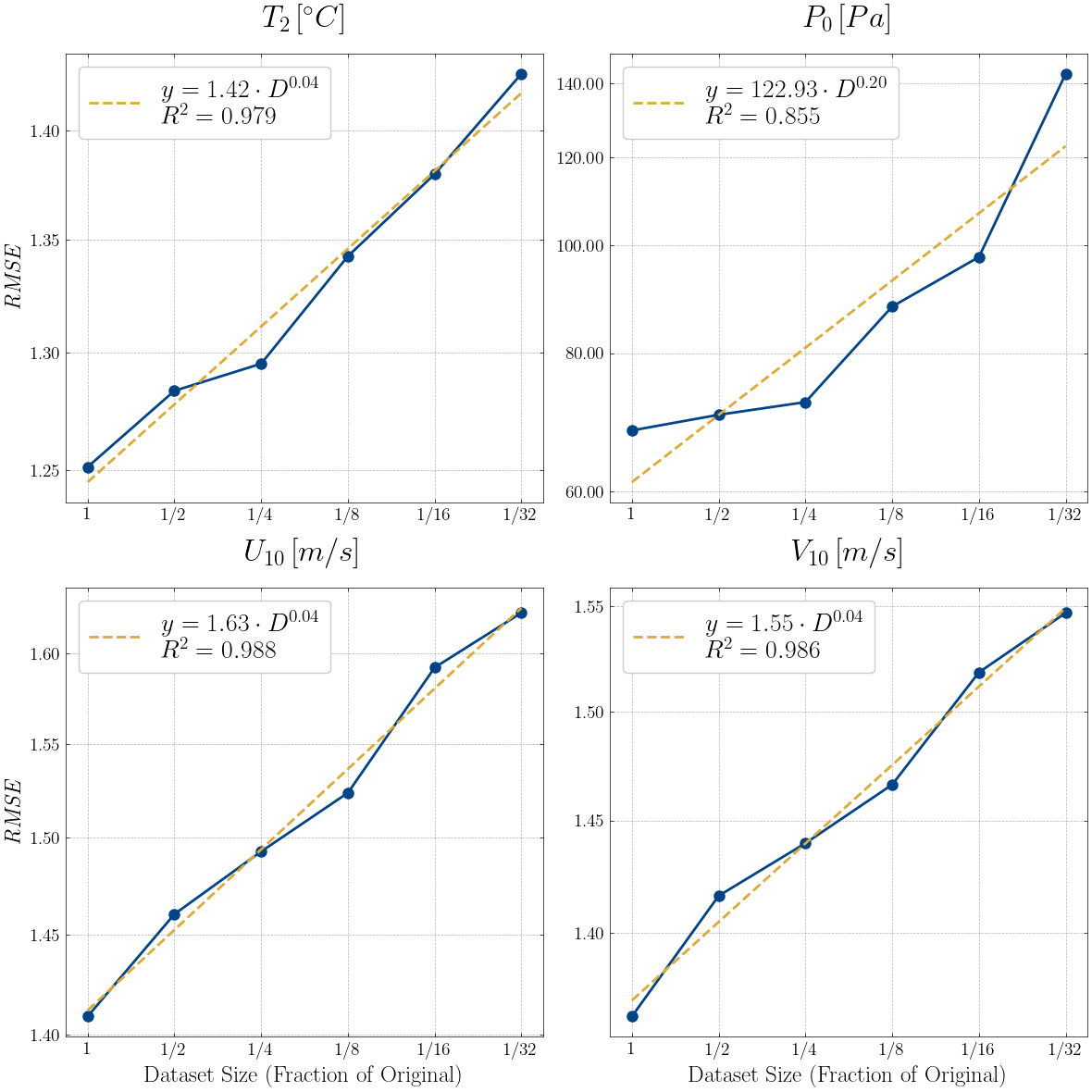}
    \caption{Scaling of decoder performance with training dataset size on WTK--US.
RMSE as a function of the fraction of training data used to train the decoder heads for near-surface temperature ($T_2$), surface pressure ($P_0$), and 10m wind components ($U_{10}, V_{10}$). The Aurora backbone is kept fixed, and only the decoder heads are trained.}
    \label{fig:scaling-laws}
\end{figure}
We evaluate how decoder performance varies with the amount of task-specific supervision by training decoder heads on progressively smaller fractions of the WTK–US training set, while keeping the Aurora backbone fixed. Figure~\ref{fig:scaling-laws} reports RMSE as a function of dataset size for near-surface temperature ($T_2$), surface pressure ($P_0$), and the 10m wind components ($u_{10}$, $v_{10}$).
For all variables, prediction error increases smoothly as the training set is reduced and follows consistent power-law trends over the evaluated data regimes. The fitted relationships indicate predictable degradation in performance as supervision decreases. Temperature and wind components exhibit similar scaling behavior, whereas surface pressure shows a stronger dependence on dataset size, indicating greater sensitivity to the amount of task-specific training data.
These observations suggest that the decoder heads leverage Aurora's pretrained representations efficiently across a range of data regimes, while additional supervision continues to yield systematic improvements.

\newpage
\section{Sample Prediction Residuals}\label{residuals}
The following plots show sample prediction residuals ($y_{\text{true}} - y_{\text{pred}}$) visualized as two-dimensional spatial fields of selected weather variables, evaluated over the CONUS region in 2010 (unseen during training) using the best-performing model.

\begin{figure}[!htbp]

    \centering
    \includegraphics[width=\linewidth]{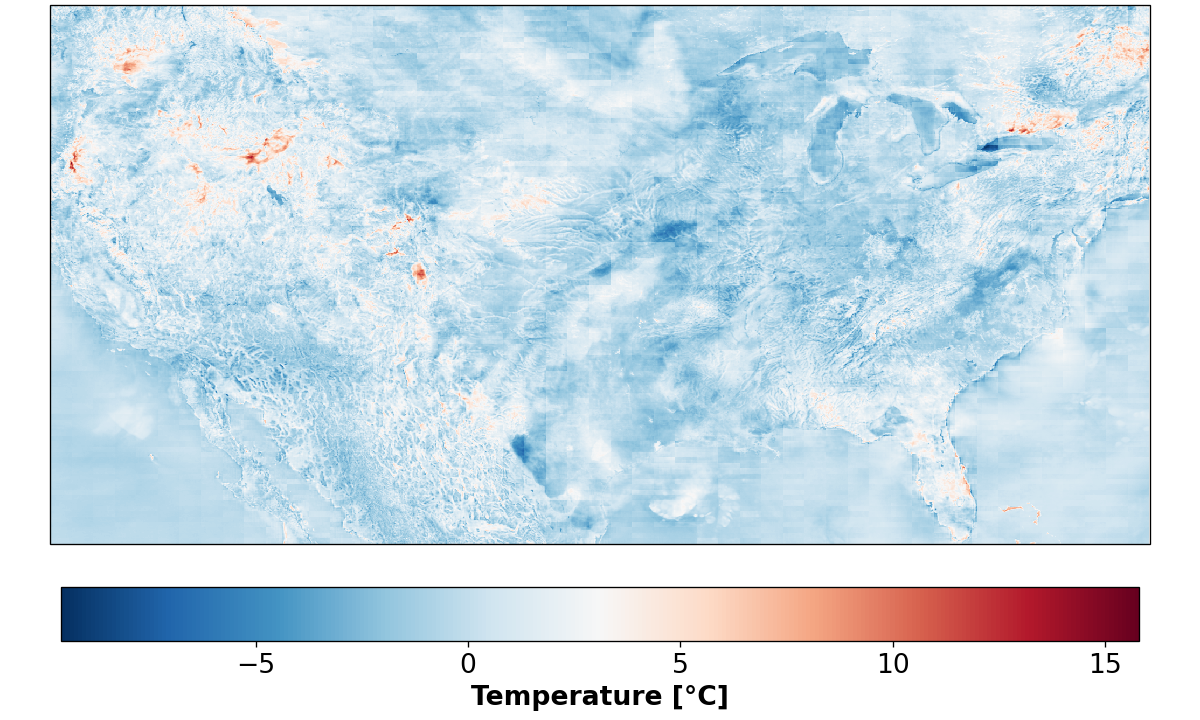}
    \caption{
        2010-02-01, residual plot, \textit{CONUS} region
        }
    \label{fig:2010-02-01_12-00_temperature_2m_inference_0.png}
\end{figure}

\begin{figure}[!htbp]

    \centering
    \includegraphics[width=\linewidth]{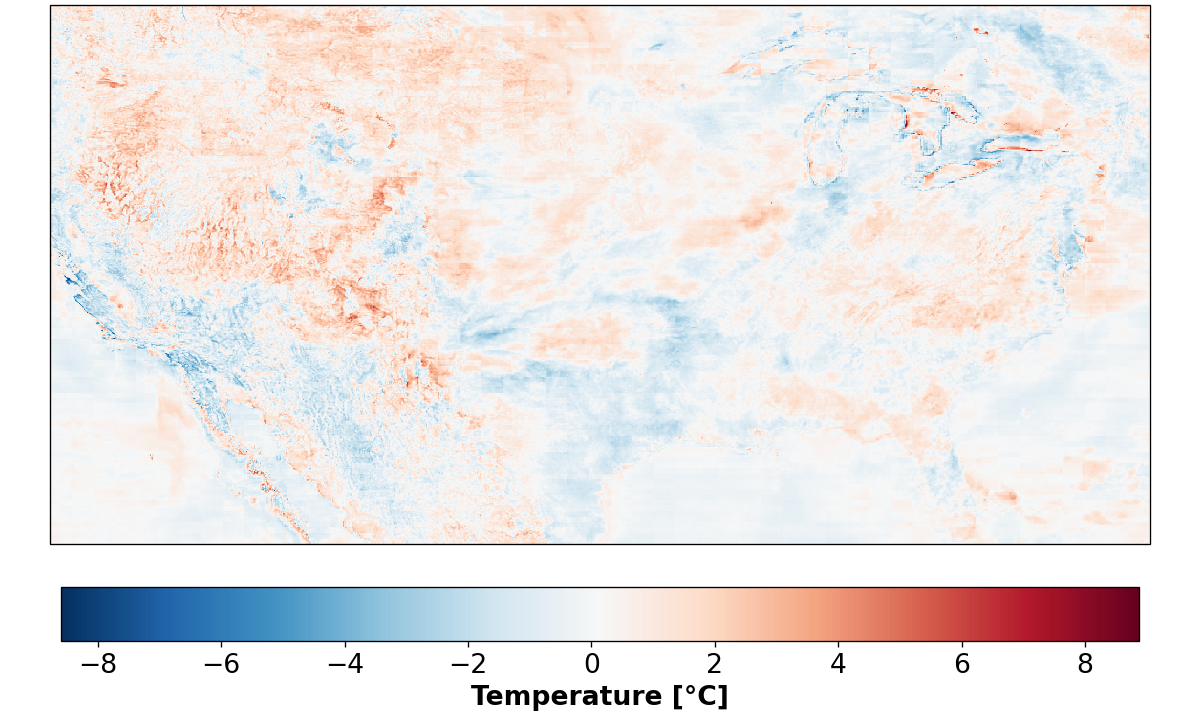}
    \caption{
        2010-06-01, residual plot, \textit{CONUS} region
        }
    \label{fig:2010-06-01_12-00_temperature_2m_inference_0.png}
\end{figure}

\begin{figure}[!htbp]

    \centering
    \includegraphics[width=\linewidth]{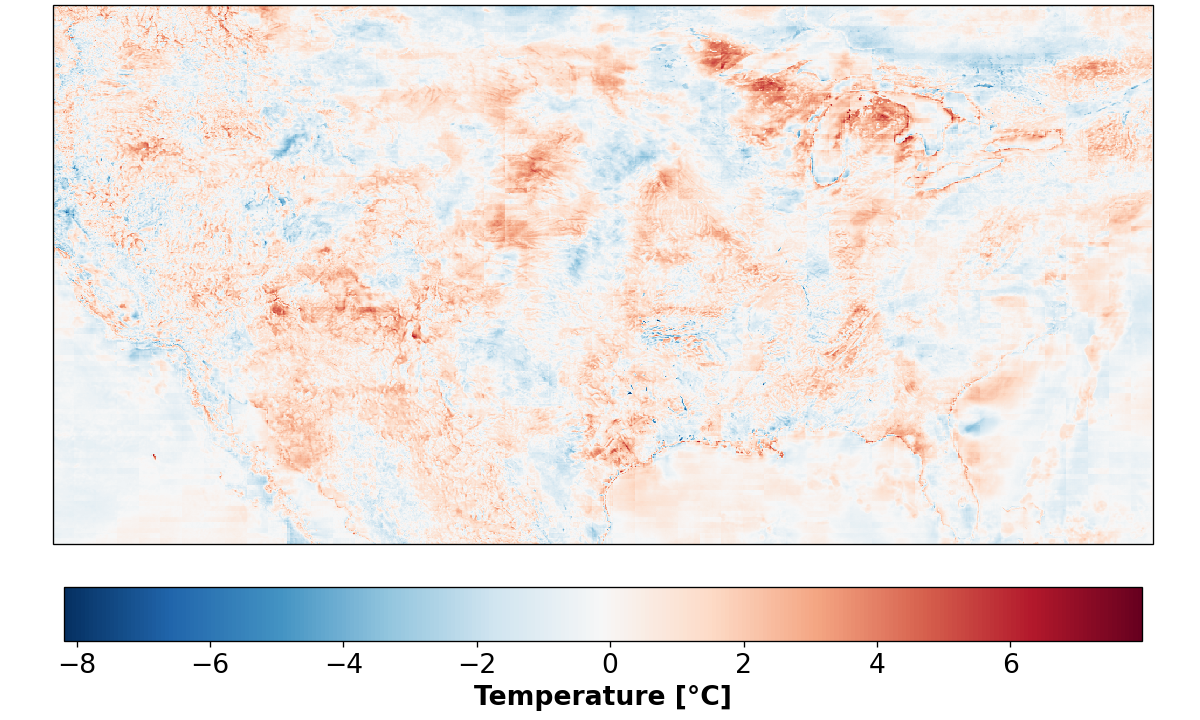}
    \caption{
        2010-10-01, residual plot, \textit{CONUS} region
        }
    \label{fig:2010-10-01_12-00_temperature_2m_inference_0.png}
\end{figure}

\begin{figure}[!htbp]

    \centering
    \includegraphics[width=\linewidth]{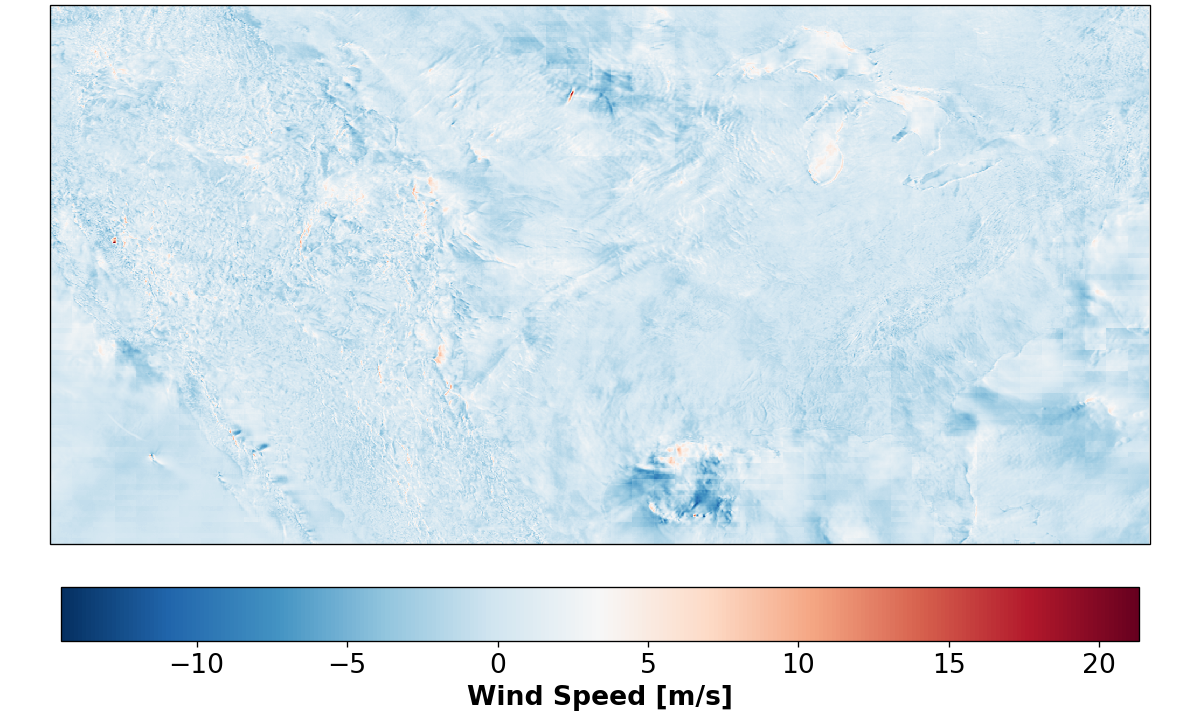}
    \caption{
        2010-02-01, residual plot, \textit{CONUS} region
        }
    \label{fig:2010-02-01_12-00_10m_u_component_of_wind_inference_0.png}
\end{figure}

\begin{figure}[!htbp]

    \centering
    \includegraphics[width=\linewidth]{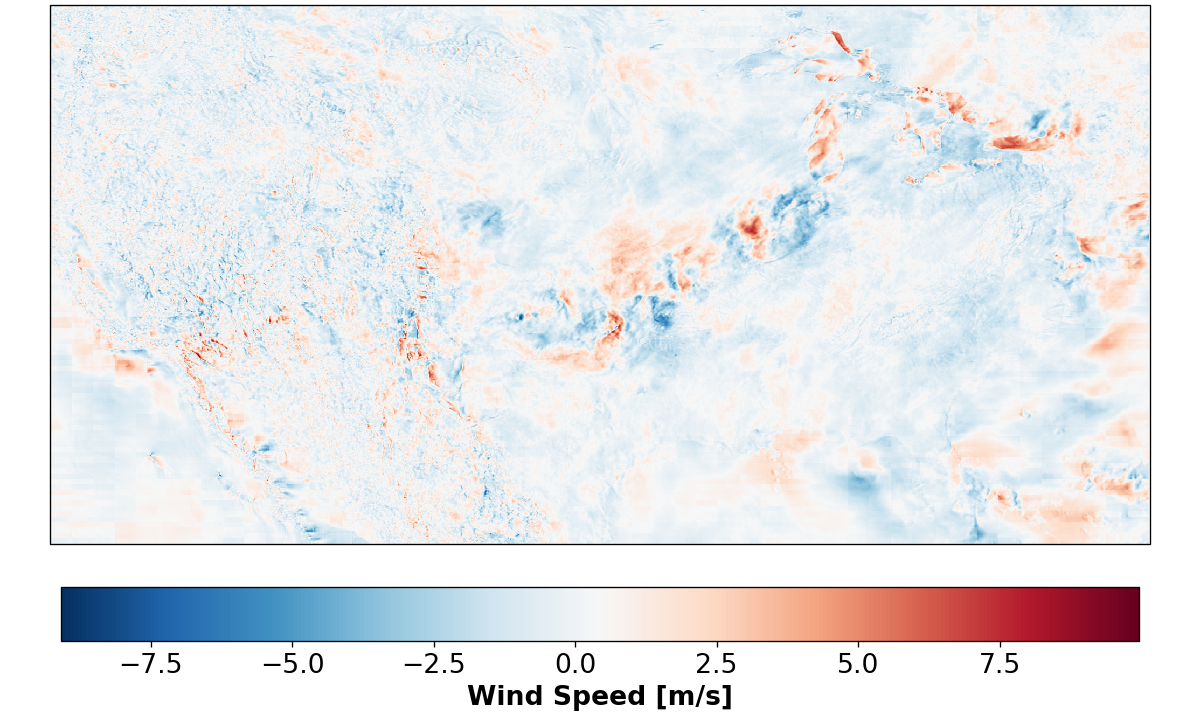}
    \caption{
        2010-06-01, residual plot, \textit{CONUS} region
        }
    \label{fig:2010-06-01_12-00_10m_u_component_of_wind_inference_0.png}
\end{figure}

\begin{figure}[!htbp]

    \centering
    \includegraphics[width=\linewidth]{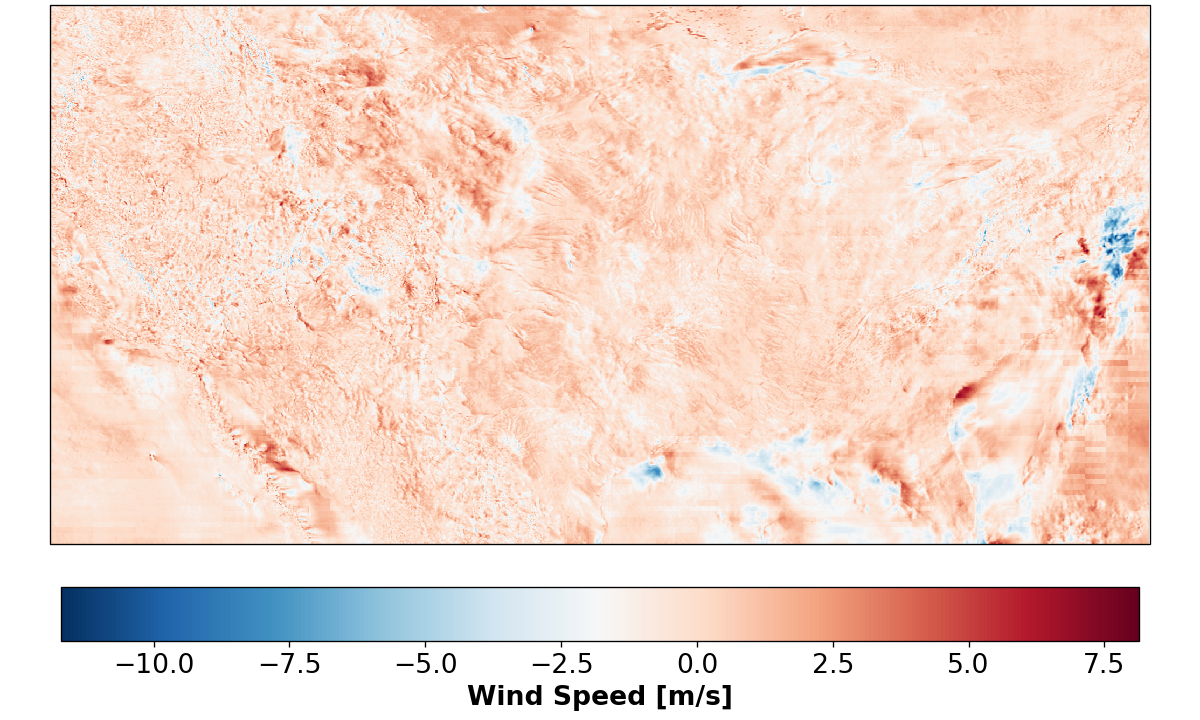}
    \caption{
        2010-10-01, residual plot, \textit{CONUS} region
        }
    \label{fig:2010-10-01_12-00_10m_u_component_of_wind.png}
\end{figure}

\end{document}